\documentclass[MS]{macro/neu_msthesis}

\usepackage{physics}
\usepackage{siunitx}
\usepackage{graphics} 
\usepackage{graphicx} 
\usepackage{epsfig} 
\usepackage{lipsum} 
\usepackage{amsmath, bm} 
\usepackage{amssymb}  
\usepackage{upgreek}
\usepackage{url}
\usepackage{breakurl}
\usepackage[breaklinks]{hyperref}
\usepackage{caption}
\usepackage{subcaption}
\usepackage{pgfplots}
\usepackage{floatrow}
\newfloatcommand{capbtabbox}{table}[][\FBwidth]

\title{Dynamic Multimodal Locomotion: A Quick Overview of Hardware and Control}

\author{Shreyansh Pitroda}

\dept{Mechanical and Industrial Engineering}

\degree{Master of Science}

\degreename{Mechanical Engineering}

\field{Mechanical Engineering}

\submitdate{August 2023}

\numberofmembers{2}

\principaladviser{Dr. Alireza Ramezani}
\firstreader{Dr. Rifat Sipahi}

\chairman{Dr. Marilyn Minus}

\dean{Jacqueline Isaacs}

\usepackage{amsfonts}			
\usepackage{times}		

\newcommand{\ifno}[1]{}

\usepackage{multirow}
\makeindex
\usepackage{makeidx}
\usepackage{acronym}

\usepackage{url}
\hypersetup{				
pdfauthor = {\authorRef},
pdftitle = {\titleRef},
pdfsubject = {\expandafter{\degreeRef} thesis submitted to Northeastern University},
pdfkeywords = {add keywords here}
pdfcreator = {LaTeX with hyperref package},
pdfproducer = {dvips + ps2pdf}}

\begin{document}

\pdfbookmark[1]{Cover}{cover}

\titlepage

\begin{frontmatter}

\pdfbookmark[1]{Table of Contents}{contents}
\tableofcontents
\listoffigures
\newpage\ssp
\listoftables



\begin{acknowledgements}

This work was made possible by all those at SiliconSynapse Lab at Northeastern University who assisted me during the process of my thesis. Prof. Alireza Ramezani, my primary adviser, has supported and believed in me from the very first day of my graduate studies. He has been instrumental in guiding and challenging me throughout my research career, in addition to giving me the opportunity to work on extremely interesting and exciting projects. I would also like to thank all of my lab mates, including Kaushik, chenghao, Adarsh, Aniket, Bibek, Xuejian, and Yizhe for their help with hardware, experiments, control design, simulation, and consistent encouragement. I would also like to thank Prof. Rifat Sipahi serving as my mechanical engineering department co-adviser and thesis reader. Finally, none of this would be possible without the support of my parents and my brother. Throughout my life, they have always pushed me to be better and been behind me every step of the way for all the things I've wanted to achieve.

\end{acknowledgements}


\begin{abstract}

Bipedal robots are a fascinating and advanced category of robots designed to mimic human form and locomotion. The development of the bipedal robots is a significant milestone in robotics. However, even the most advanced bipedal robots are susceptible to changes in terrain, obstacle negotiation, payload, and weight distribution, and the ability to recover after stumbles. These problems can be circumvented by introducing thrusters. Thrusters will allow the robot to stabilize on various uneven terrain. The robot can easily avoid obstacles and will be able to recover after stumbling. Harpy is a bipedal robot that has 6 joints and 2 thrusters and serves as a hardware platform for implementing advanced control algorithms. This thesis explores manufacturing harpy hardware such that the overall system can be lightweight and strong. Also, it goes through simulation results to show thruster-assisted walking, and at last, it shows firmware and communication network development which is implemented on actual hardware.

\end{abstract}

\end{frontmatter}


\pagestyle{headings}


\chapter{Introduction}
\label{chap:intro}

Bipedal robots are machines that are designed to mimic human-like locomotion to transverse challenging terrains. The bipedal robots can access areas with narrow passages, climb stairs\cite{conference:stairclimb}, hop\cite{conference:hopping}, and perform back-flips. However, even the most robust bipedal systems are not capable of recovering from a slip or disturbance once a certain threshold has been reached. A bipedal robot has severe limitations due to which it cannot be operated in the outside environment where terrain is too uncertain. By introducing thrusters to a biped robot, these challenges can be negotiated by allowing the robot to jump over challenging terrain and by assisting in stabilization.

\begin{figure}[h]
  \centering
      \begin{subfigure}[b]{0.5\textwidth}
        \centering
        \includegraphics[width=0.90\textwidth]{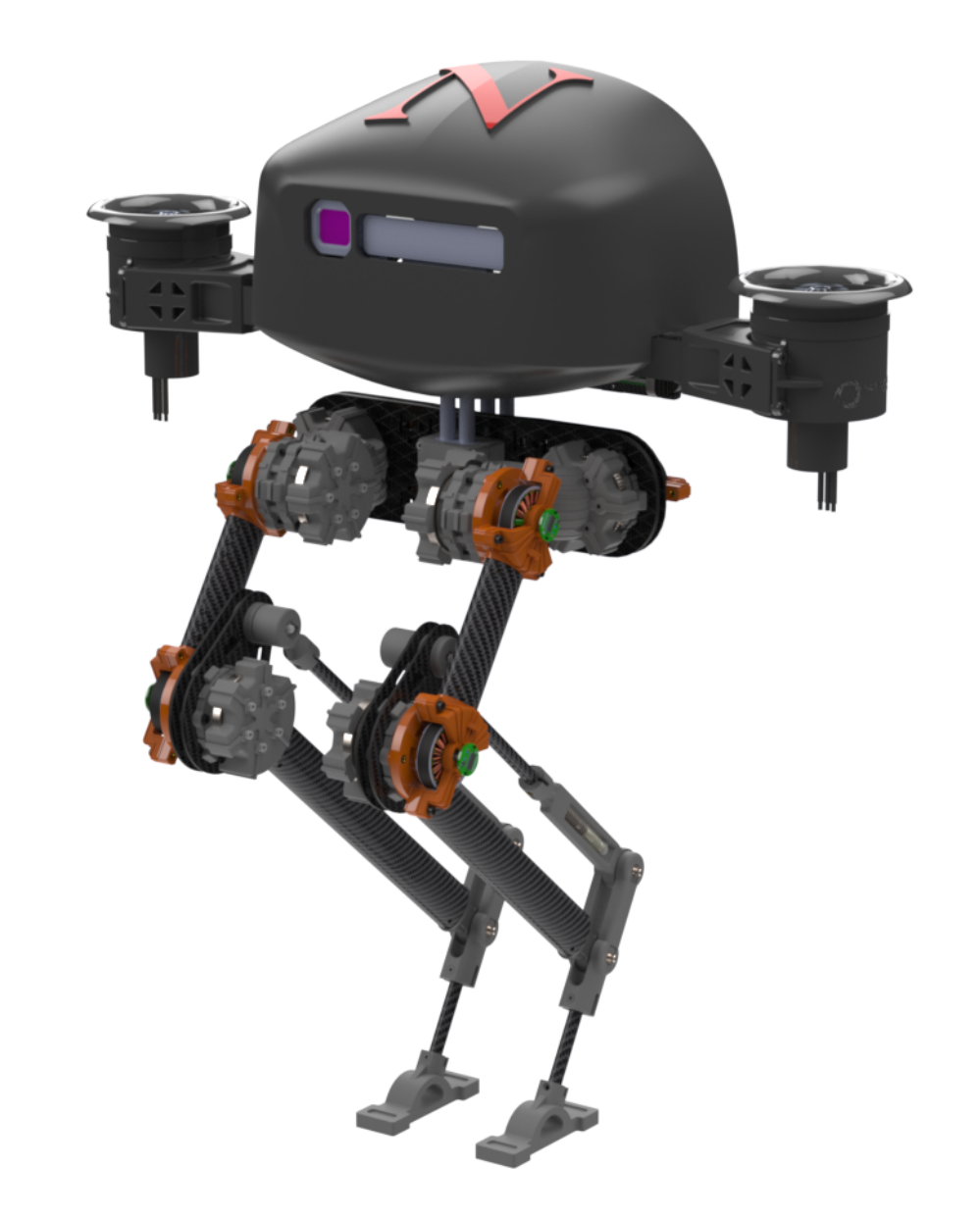}
        \caption{Render of Harpy}
        \label{fig:HarpyRender}
     \end{subfigure}
     \hfill
     \begin{subfigure}[b]{0.45\textwidth}
        \centering
        \includegraphics[width=1\textwidth]{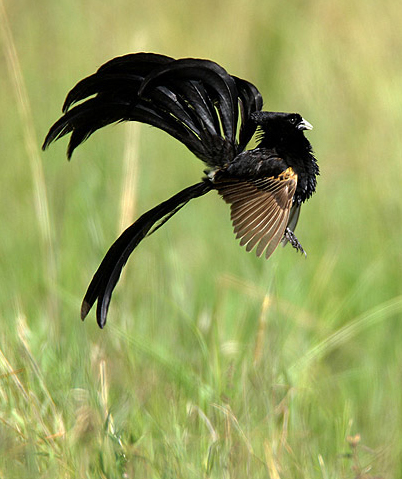}
        \caption{Widowbird mid-jump \cite{web:widowbird}}
        \label{fig:widowbird}
     \end{subfigure}
     \hfill
    \caption[Render of Harpy and Jackson's Widowbird mid-jump]{Harpy draws design inspiration from nature, and particularly from the Jackson's Widowbird}
    \label{fig:HarpyComparison}
\end{figure}

This thesis introduces Harpy, a bipedal robot incorporating thrusters (depicted in Fig. \ref{fig:HarpyRender}). The incorporation of thrusters brings about unprecedented capabilities in handling disturbances and recovering from them, thereby facilitating navigation across challenging terrains. The robot's remarkable thrust-to-weight ratio further empowers it to surmount substantial obstacles and traverse difficult landscapes. The design inspiration for Harpy is drawn from the Widowbird, specifically Jackson's Widowbird, a diminutive avian species indigenous to Kenya and Tanzania. The Widowbird employs both its wings and legs to execute remarkably high vertical leaps that surpass its own body length, as a crucial aspect of its mating behavior (illustrated in Fig. \ref{fig:widowbird}). This behavior drives male Widowbirds to engage in competitive jumping to achieve greater heights and frequencies. The avian species possess the ability to swiftly and agilely move using its legs, as well as to fly, all while maintaining a lightweight and energetically efficient build. Given the impracticality of replicating avian wings on a legged robot, the Harpy robot instead integrates thrusters to achieve similar functionalities.

\section{Legged Robotic Systems}
\label{chap:intro:leggedRobots}

Bipedal robots like Boston Dynamics' Atlas \cite{paper:ATLAS}, Agility Robotics' Cassie \cite{conference:Cassie}, Ascento \cite{conference:ascento}, ANYmal \cite{conference:ANYmal} and OSU's ATRIAS \cite{paper:ATRIAS} have demonstrated robust disturbance rejection and adept navigation in unfamiliar settings. However, a critical limitation is their inability to recover once a fall is initiated. This shortcoming presents significant challenges in scenarios such as search and rescue, disaster response, and tasks conducted in hazardous environments. The absence of human intervention nearby to restore the robot's functionality amplifies this issue. While certain humanoid robots, including Atlas, can regain an upright posture post-fall using their arms, these robots still face constraints when confronted with rugged terrains.

\subsection{Thruster-assisted Legged Robots}

\begin{figure}[h]
  \centering
      \begin{subfigure}[b]{0.45\textwidth}
        \centering
        \includegraphics[width=\textwidth]{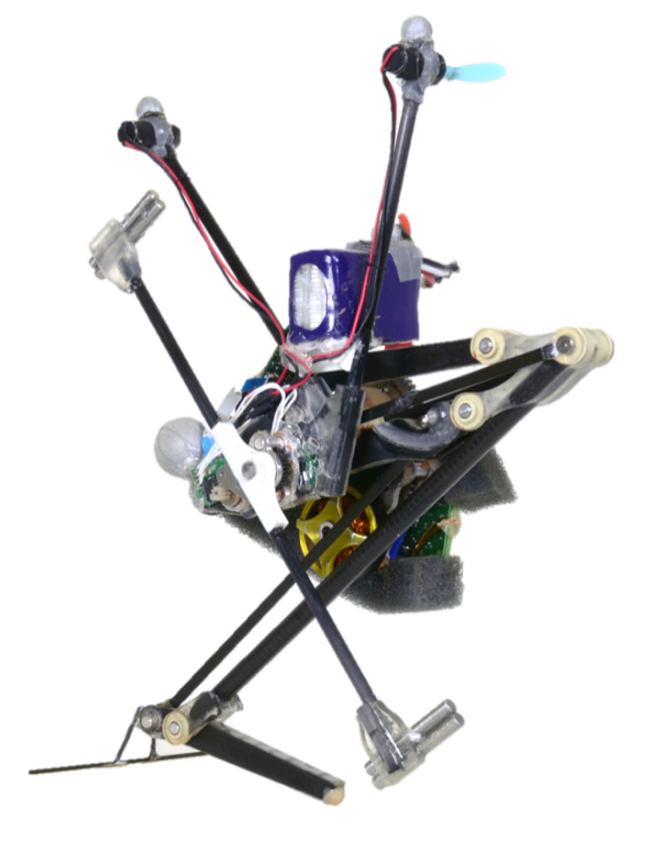}
        \caption{Salto-1P \cite{salto1p}}
     \end{subfigure}
     \hfill
     \begin{subfigure}[b]{0.45\textwidth}
        \centering
        \includegraphics[width=\textwidth]{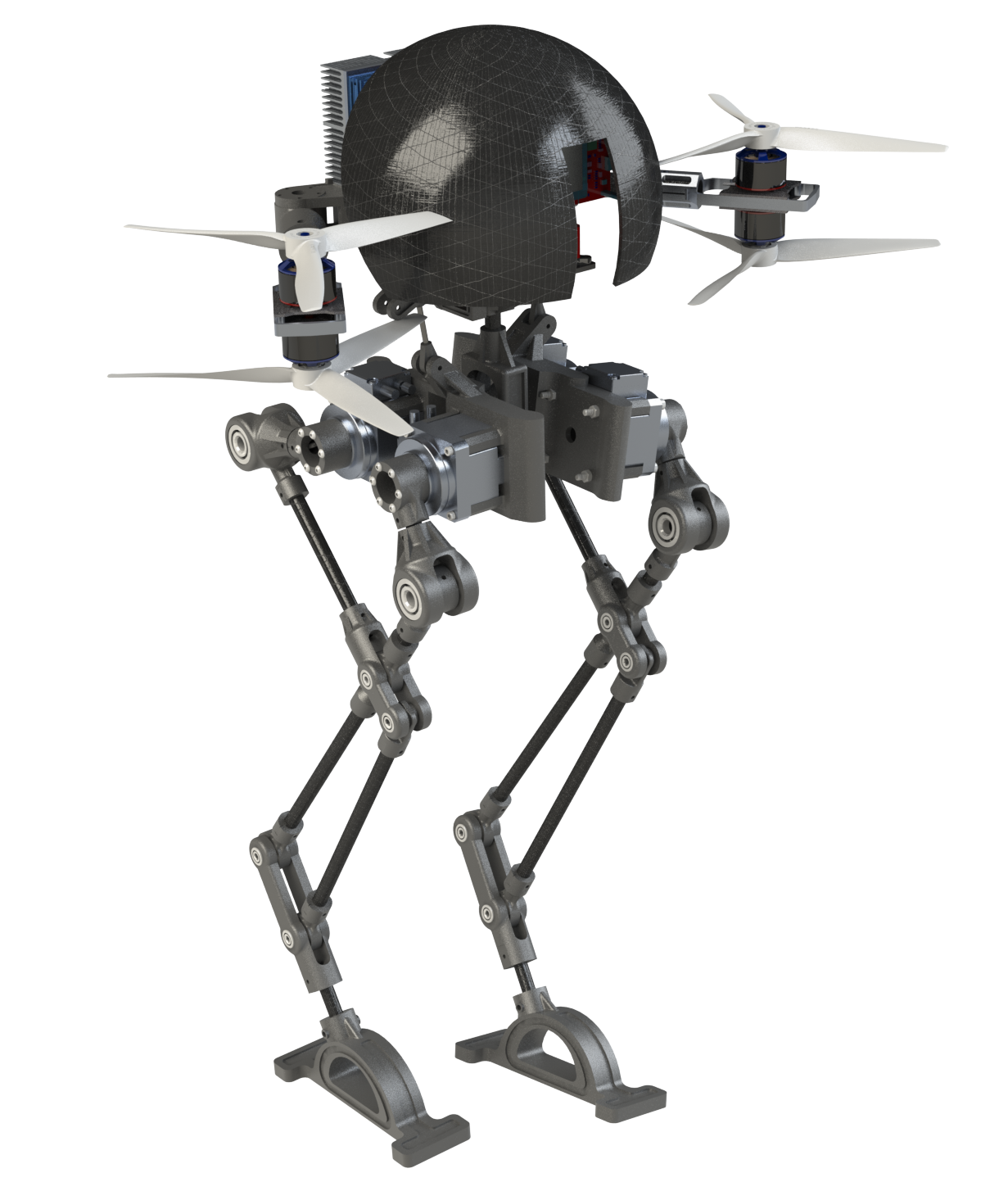}
        \caption{Leonardo \cite{article:Leonardo}}
     \end{subfigure}
     \hfill
    \caption[Thruster-assisted legged robots]{Thruster-assisted legged robots}
    \label{fig:saltoLeo}
\end{figure}

At the time of writing, Salto-1P \cite{salto1p} and Leonardo \cite{article:Leonardo}, as depicted in Fig.~\ref{fig:saltoLeo}, are two additional legged robots that showcase thruster integration. Salto-1P, an innovative monopedal hopping robot developed at UC Berkeley, features a geared brushless motor to actuate its leg and a flywheel for orientation control. Impressively, despite a relatively short leg length of 144 mm, it achieves jumps of up to 1.25 m. The robot employs micro-brushed DC motors with propellers to enhance stability during mid-air maneuvers. It capitalizes on its energy-dense leg and a series-elastic spring mechanism to execute substantial leaps. Notably, Salto-1P's predecessor, Salto, lacked thrusters and could only perform a limited number of hops before necessitating the combined control of the flywheel and thrusters.

Caltech and Northeastern's Leonardo, on the other hand, has successfully incorporated thrusters into a bipedal legged robot. This robot demonstrates stabilized walking gaits, single-leg balancing, and spinning maneuvers, as showcased in its operation. Also, Leonardo is able to perform flying to get down the stairs, slackline walking, and skating.

It's important to note that, as of the time of writing, neither of these robots has used a combination of thrusters and legs in parallel like high jumps and slope walking. Instead, their primary application has been to enhance stability during various locomotion tasks.

While not involving thruster assistance, a similar conceptual approach integrated into legged robots involves utilizing balloon buoyancy to introduce a passive force opposing gravity. UCLA's Ballu \cite{conference:ballu}, for instance, adopts this strategy by incorporating a sizable helium balloon. Ballu is designed as an exceptionally lightweight bipedal robot, featuring simple string-actuated legs with only one degree of freedom (DOF). Despite its capabilities for walking, hopping, and turning, its precise control is challenging due to the absence of active actuation and the inherent unpredictability associated with the relatively flexible and large balloon.

Robots employing balloons benefit from notable energy efficiency; however, they face limitations when it comes to reliable outdoor operation. Even a mild breeze can propel the robot away, given the substantial surface area of the balloon. Furthermore, this approach becomes less feasible for larger and more complex robots with heavier payloads, as it would require disproportionately large balloons, rendering them impractical.

\subsection{Husky Carbon}

\begin{figure}[h]
  \centering
    \includegraphics[width=0.7\textwidth]{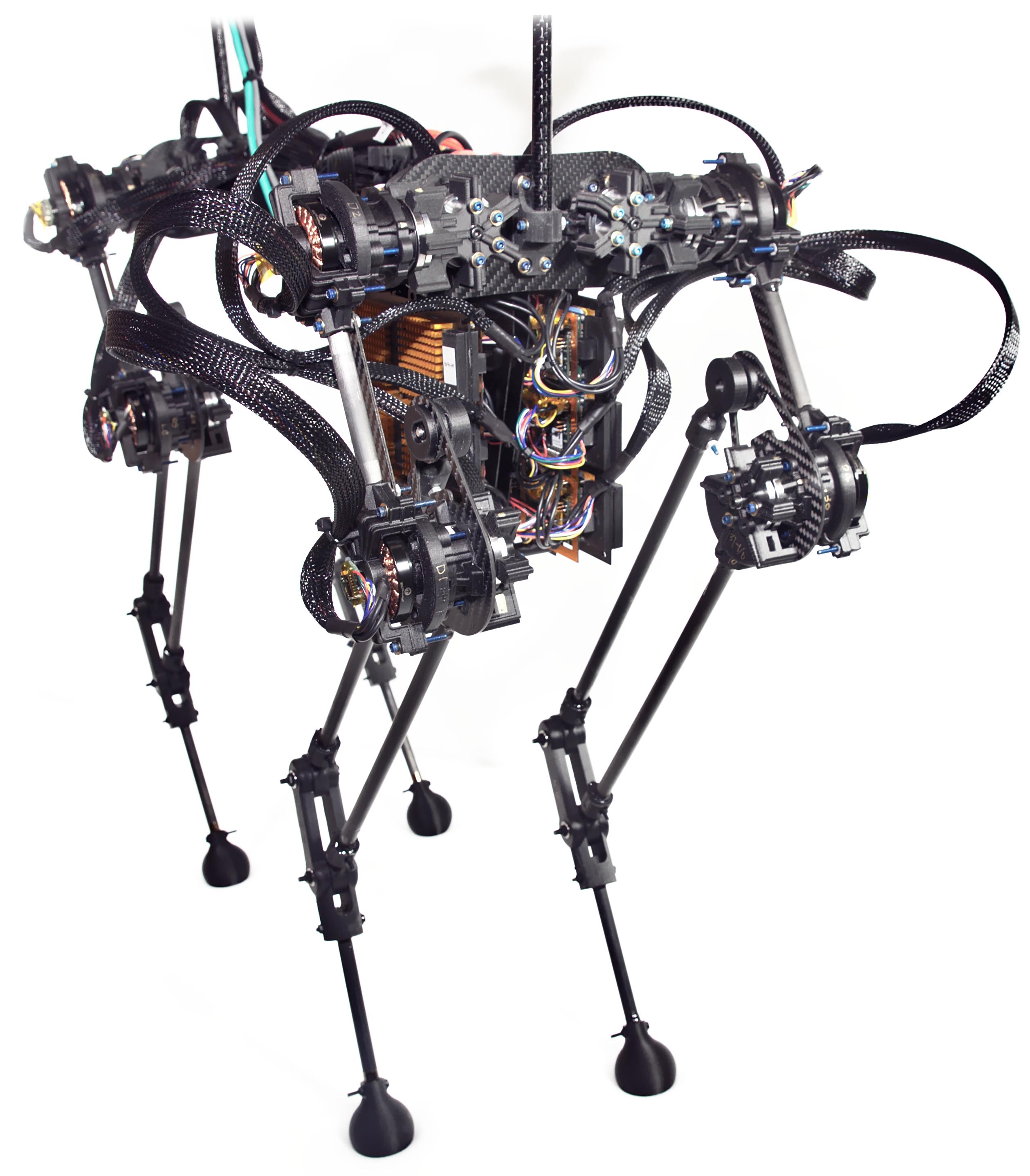}
  \caption[Husky Carbon]{Husky Carbon (image courtesy of Pravin Dangol)}
  \label{fig:Husky}
\end{figure}

Prior to Harpy's development, the SiliconSynapse Lab at Northeastern University introduced the Husky Carbon\cite{paper:Husky}\cite{unknown:huskypaperMPC} (depicted in Fig \ref{fig:Husky}), a remarkably lightweight quadrupedal-legged robot. One of its distinctive features is its capacity to transform into a quadrotor configuration, a configuration aimed at exploring the possibilities of hybrid legged-aerial locomotion. Notably, this approach diverges from Harpy's hybrid locomotion strategy, as Husky Carbon employs two distinct modes rather than merging thrusters and legs to attain heightened stability and performance.

Harpy not only builds upon Husky's\cite{paper:Huskygenerative} innovations but also refines them. Specifically, Harpy enhances design elements initially validated on the Husky platform. This includes the strategic incorporation of components within 3D printed structures and the utilization of carbon fiber tubes and plates for constructing leg and body frameworks. These choices serve to minimize overall weight while simultaneously maximizing stiffness and robustness. Furthermore, Harpy leverages components from Husky Carbon's leg joint actuators to simplify the design and reduce costs associated with Harpy's thruster actuators.

\section{Control Advantages}

\begin{figure}[h]
  \centering
    \includegraphics[width=1\textwidth]{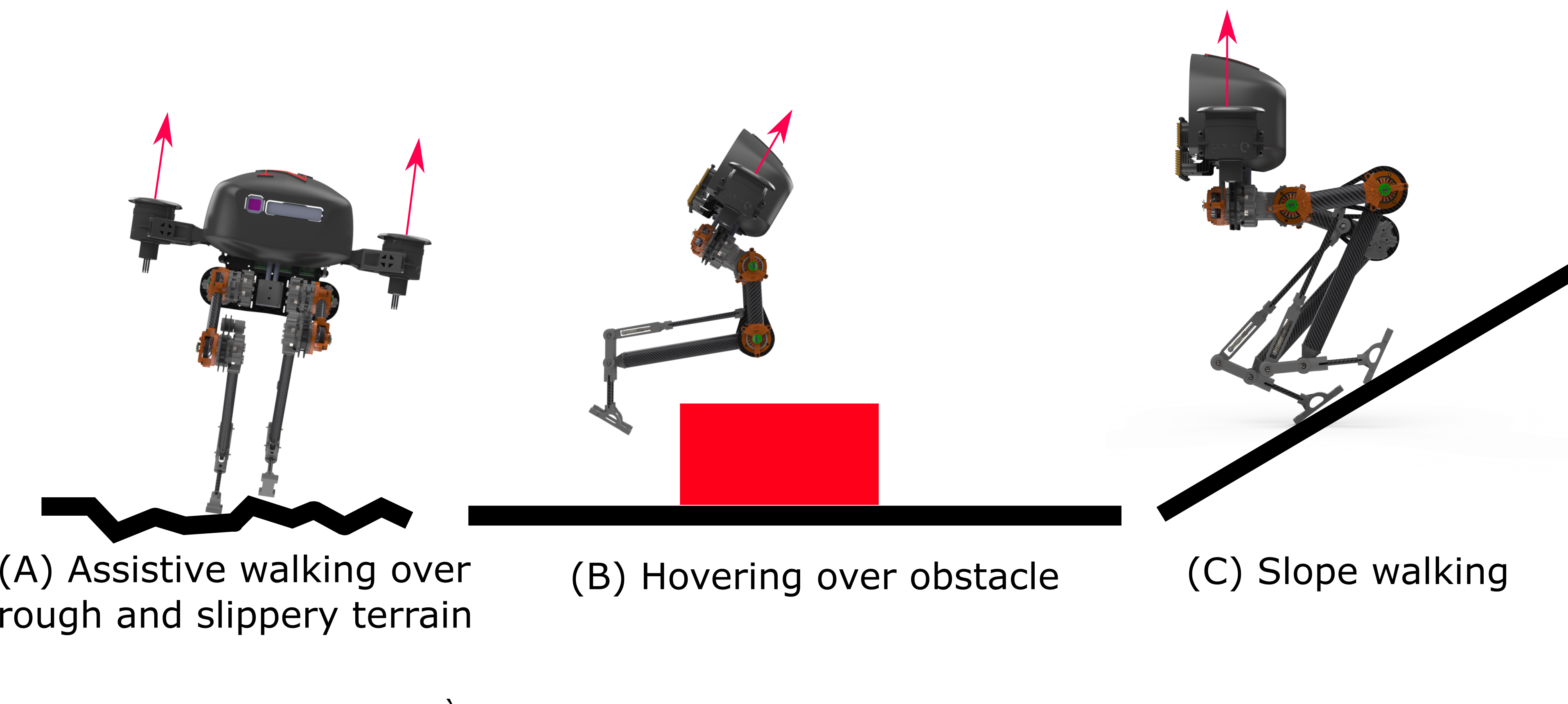}
  \caption[Husky Carbon]{harpy performing different tasks}
  \label{fig:harpy performing task}
\end{figure}

Since the thruster stabilizes the frontal dynamics of the robot, it greatly simplifies the control problem into a quasi-2D legged system. Due to this stabilization, we can operate Harpy in different environments. By using thruster upward force, the harpy can walk on a steep slope as shown in (Fig \ref{fig:harpy performing task}a). Next, we can use Harpy for performing search and rescue missions since can walk and fly over larger obstacles as shown in (Fig \ref{fig:harpy performing task}b). lastly, due to frontal dynamics stabilization, it allows us to use Harpy for space exploration where the terrain is uneven as shown in (Fig \ref{fig:harpy performing task}c)

\section{Project Goals}

\begin{figure}[h]
  \centering
    \includegraphics[width=0.7\textwidth]{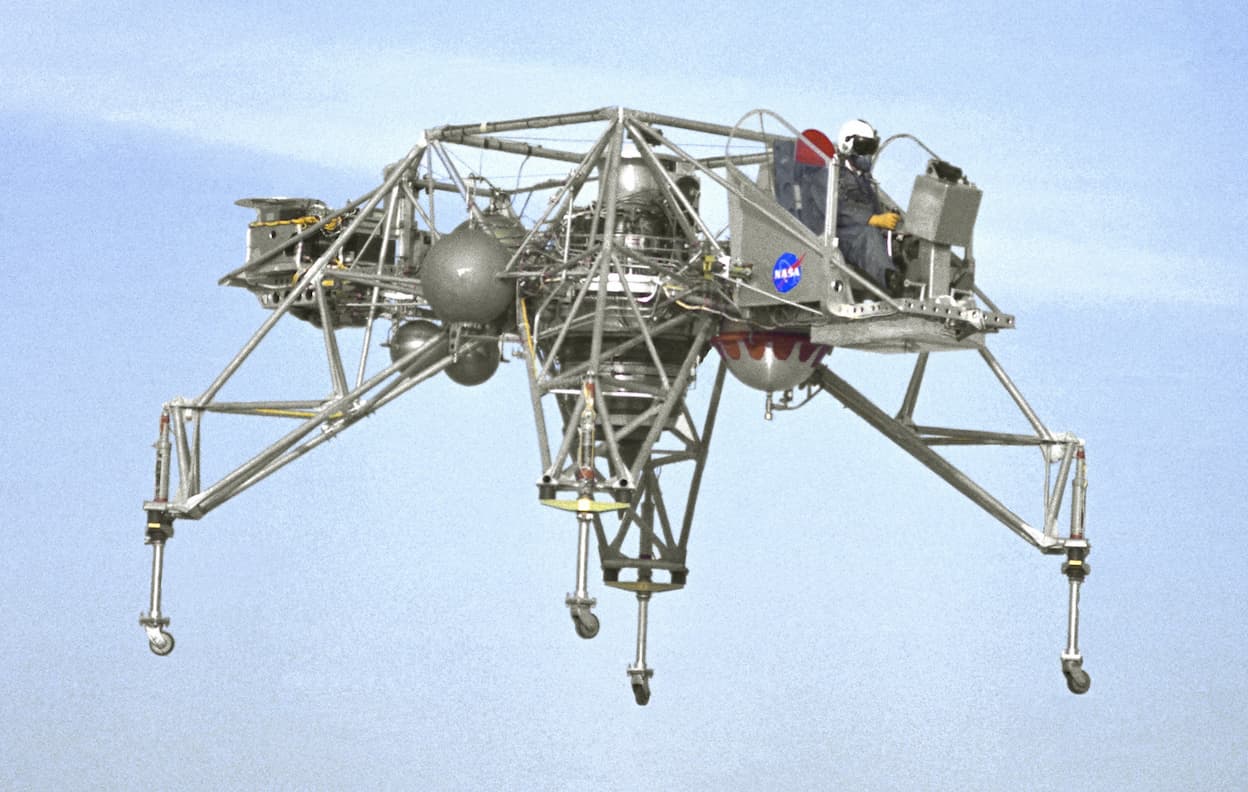}
  \caption[Lunar lander research vehicle]{Lunar lander research vehicle \cite{web:LLRV}}
  \label{fig:llrv}
\end{figure}

Harpy's\cite{unknown:ManifoldHarpy}\cite{unknown:RoughterrainHarpy}\cite{unknown:GRFHarpy}\cite{article:Thrusterassistedcontrol} primary mission is to serve as a robust hardware foundation for the advancement of control strategies in the domain of thruster-assisted legged locomotion. This encompasses the refinement of stability and equilibrium during walking through the integration of thrusters, as well as the capability to surmount obstacles via jumping. Moreover, Harpy is strategically designed to act as a versatile platform for the comprehensive study of bipedal-legged locomotion under low-gravity conditions. This approach is reminiscent of the Lunar Lander Research Vehicle's (depicted in Fig.~\ref{fig:llrv}) role in investigating human piloting techniques for landing the Apollo Lunar Module on the Moon, and its subsequent application in training Lunar Module pilots. Notably, the Lunar Lander Research Vehicle achieved this by employing a jet engine mounted on a gimbal to counteract $83$ percent of the aircraft's weight, effectively simulating lunar gravity conditions. In a similar vein, Harpy leverages its pair of actuated thrusters to replicate this effect, facilitating research pertaining to bipedal robots optimized for space exploration endeavors.

In addition to these goals, the main constraints driving the electromechanical design of the system are as follows:

\begin{enumerate}

  \item \textbf{Low Weight:} A pivotal requirement mandates Harpy to achieve a thrust-to-weight ratio exceeding unity, all while upholding a notable degree of robustness and stiffness. This feat is accomplished through strategic adoption of composite 3D printing and carbon fiber components, thereby mitigating the integration of heavy materials such as metals whenever feasible.

  \item \textbf{Impact Resistance: } Given Harpy's role as a jumping legged robot, its legs are engineered to endure and effectively absorb the impact generated from substantial free falls. To address this challenge, passive shock absorbers are seamlessly integrated into the ankle joints, thereby reducing the torque exerted on the leg actuators. Furthermore, these shock absorbers have the capacity to store and subsequently release energy for subsequent jumping maneuvers.

  \item \textbf{Modularity:} An essential aspect of Harpy's design philosophy is its modularity, a characteristic of paramount importance for a research-oriented hardware platform. While modularity often introduces additional hardware, complexity, and weight, Harpy achieves a well-calibrated balance. Notably, despite the permanent integration of certain actuators within the knee and frontal hip joints, their design enables straightforward removal and replacement of harmonic drive components without necessitating the destruction of 3D-printed housing. This modular approach extends to motor housing as well, offering the flexibility to detach it from any joint for the purpose of replacing or modifying various components.
  
\end{enumerate}



\chapter{Northeastern's Harpy platform}
\label{chap: Northeastern's Harpy platform}

This chapter provides an overview of Northeastern's harpy hardware platform. Harpy consists of two main subsystems namely Legged assembly and Aerial assembly. The design of the harpy's legged assembly was done by a previous MS and PhD student and preliminary tests were performed by prototyping one leg. This chapter gives an overview of harpy's legged and aerial assembly.

\begin{figure}[h]
  \centering
    \includegraphics[width=1\textwidth]{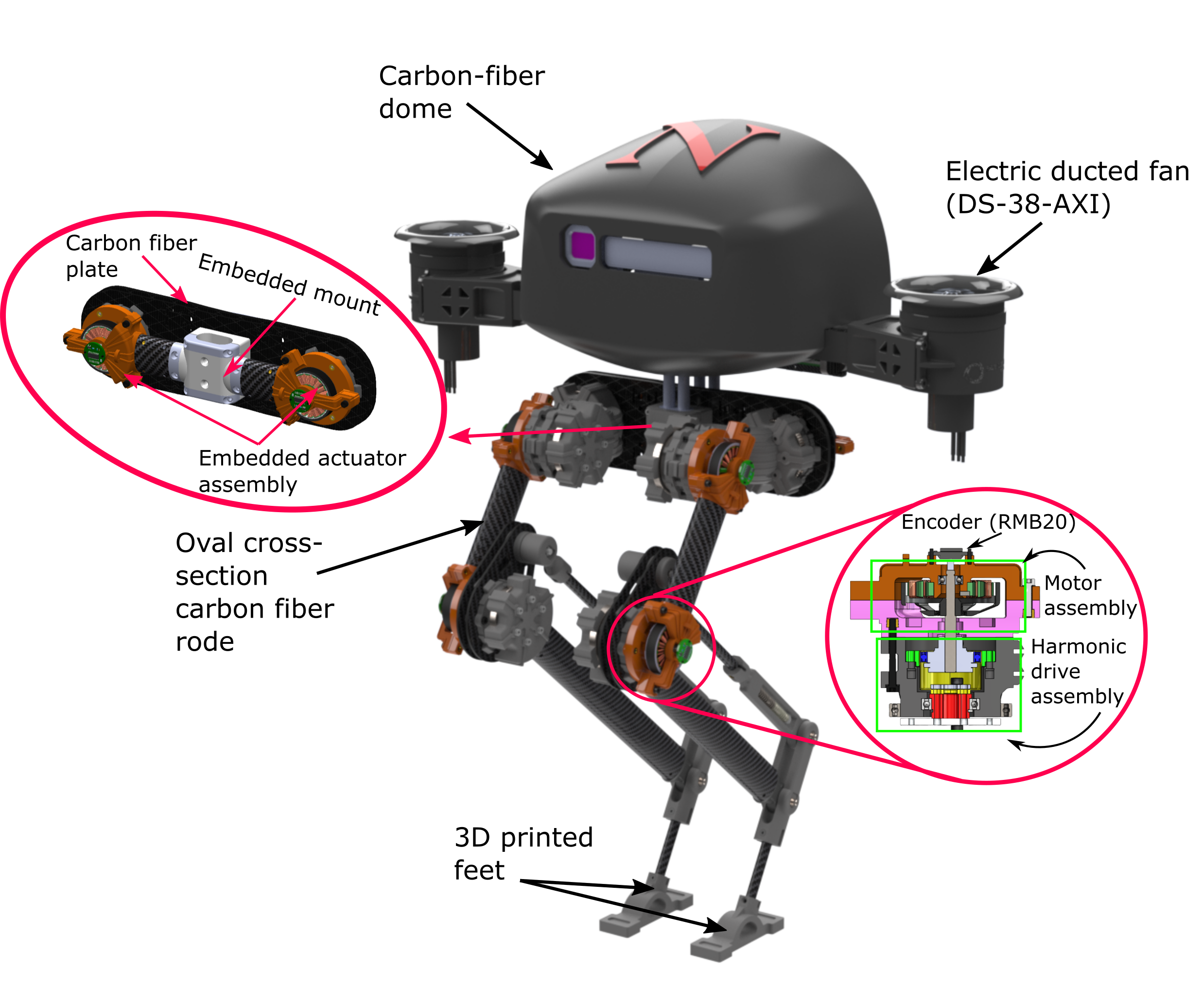}
  \caption[Harpy model]{Harpy Model} 
  \label{fig:Harpy Model}
\end{figure}

Harpy's design was developed in silicon synapse, shown in fig (\ref{fig:Harpy Model}), Harpy is 30 inches tall, 24 inches wide and weighs about $4$kg. Harpy's body is designed to be energy dense, lightweight, and robust so it can perform jumps, absorb impacts, and have inertia closer to the foot to allow easier foot placement. Harpy's leg design is a series actuated pantagraph design due to its mechanical advantage and efficient use of actuator torque. Each leg has three degrees of freedom provided through Hip frontal joints (HFJ), Hip Sagittal joints (HSJ), and Knee joints (KJ). Harpy's thruster mount is designed using a composite structure to have a higher strength-to-weight ratio.

\section{Design overview - Legged assembly}

\subsection{Pelvis block assembly}

The pelvis block is a critical component in bipedal robots as it serves as the central structure connecting the robot's leg, providing stability, balance and supporting the upper aerial assembly. The pelvis block is responsible for mitigating force exerted on the leg assembly during walking, jumping, and other locomotion. Thus, the pelvis block must be constructed from lightweight yet durable material that can withstand the forces. Thus, Harpy's pelvis block as shown in fig(\ref{fig:pelvis block}) is designed to have two pelvis carbon fiber plates that are placed 8mm apart. These plates ensure that there is no deflection in the transverse direction. These plates are embedded into the frontal actuator and to further reduce the deflection in the frontal direction, there are two oval carbon fiber rod that are connected to the central connector and frontal motor assembly. 

\subsection{Leg assembly overview}

The selection of a pantograph, series actuated leg design for Harpy stems from its advantageous mechanical characteristics: efficiency in leveraging actuator torque, lightweight composition, and the avoidance of a singular configuration and knee inversion. Oval cross-section carbon fiber tubes are employed to optimize the balance between stiffness, strength, and weight within the link structure.

Contrasted with circular cross-section tubes that provide uniform bending resistance across all directions, the application of oval tubes focuses on maximizing stiffness within the robot's sagittal plane. This plane encounters the greatest stress during walking and jumping. Consequently, this design choice not only reduces overall weight but also conserves space occupied by the leg links, particularly when compared to circular tubes with a diameter equivalent to the major diameter of the oval tubes. To ensure structural integrity, all couplings of carbon fiber tubes are secured through a criss-cross pattern of fasteners, effectively constraining rotation along all axes.

Incorporating series elastic actuation offers several key benefits. It excels in force control and exhibits heightened robustness when confronted with external disturbances. The elastic component serves as a low-pass filter, effectively attenuating high-frequency disturbances, resulting in smoother and more stable responses. Additionally, cyclic motions such as trotting, walking, and jumping enable the elastic element to store and subsequently release energy. This not only contributes to smoother motion but also reduces overall energy consumption.

\subsubsection{Actuator}

\begin{figure}[h]
  \centering
    \includegraphics[width=1\textwidth]{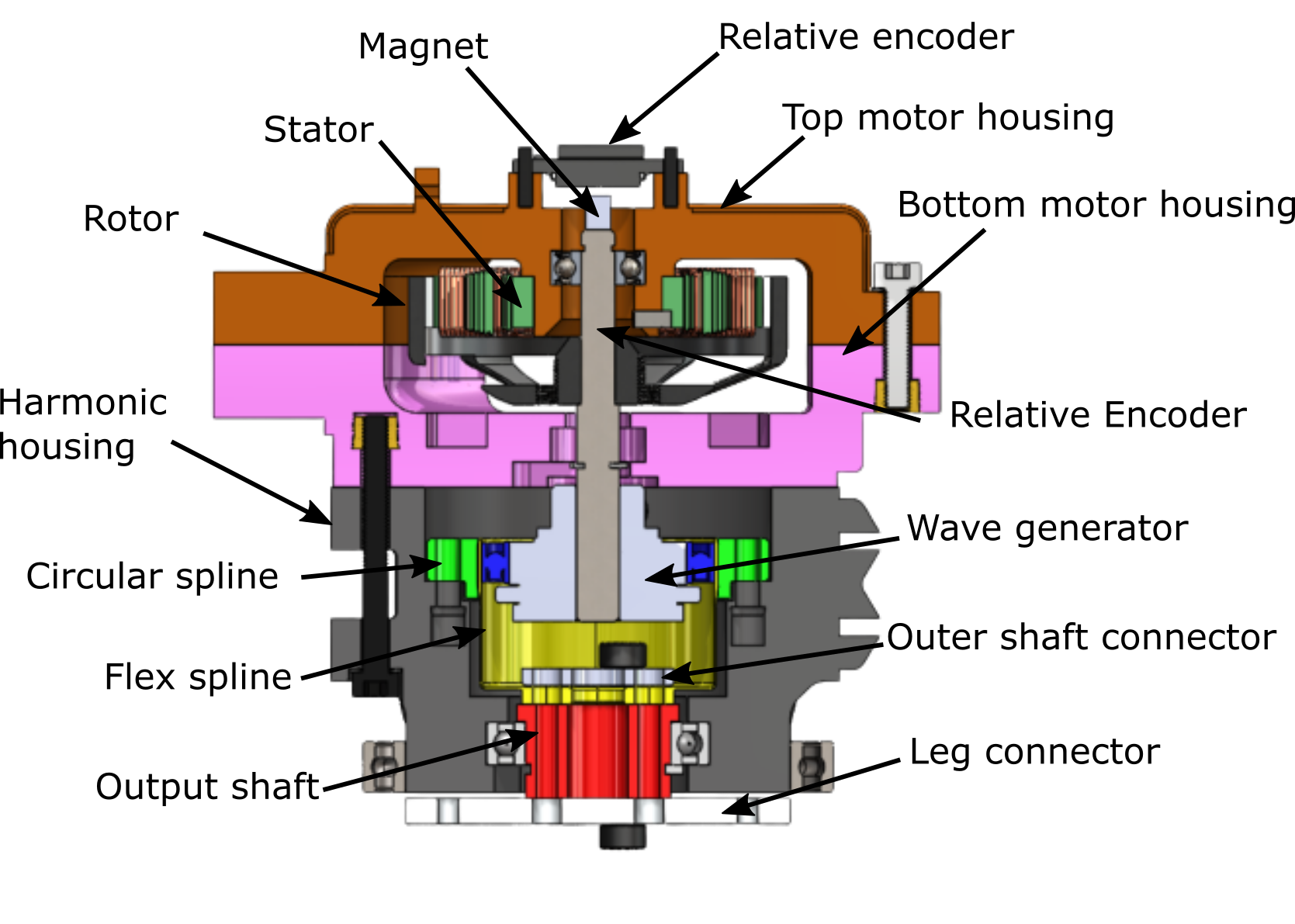}
  \caption[Harpy model]{Harpy's actuator sectional view} 
  \label{fig:Actuator sectional viewl}
\end{figure}

Legged robot joint actuators in general must be extremely power dense with high transparency, high efficiency, and minimal backlash. The importance of the joint lies in its fundamental role in enabling robots to mimic the bio-inspired movement and accomplish a wide variety of tasks with precision and adaptability.The methodology behind selecting components and materials for this robot’s leg joint actuators are discussed in this section. 

In the design of a Harpy's actuator transmission, several critical qualities must be considered, including mechanical transparency, torque density, back-drivability, and low backlash. High transparency actuators allow efficient energy flow from the motor to the output shaft, typically achieved with low gear ratios. Torque density is essential for achieving a high overall thrust-to-weight ratio in the robot, while back-drivability ensures better impact absorption and high transparency. Low backlash is crucial to minimize uncertainty in the robot's leg position. Direct drive systems offer high efficiency, precise control, and low backlash but may be limited by the size and weight of the motor. Planetary gear systems provide high torque density and can handle significant loads but might suffer from increased friction and complexity. Harmonic drives combine features from planetary gears and flexible components, offering high reduction ratios with minimal backlash and high torque transmission. They also offer modularity in robot actuators due to their interchangeable gear ratios. The selection of the transmission type affects the input torque developed by the motor's electromagnetic torque, which, in turn, determines the output torque for each joint actuator. Considering these factors, a harmonious drive system was chosen to strike a balance between efficiency, torque density, and low backlash for the legged robot's actuator design.

The decision to opt for a harmonic drive was driven by its array of advantageous attributes, as previously mentioned. Specifically, we chose the CSF-11-30-2A-R model due to its capability to deliver elevated torque output. Importantly, this selection facilitates the attainment of both swift joint movement and exceptional back-drivability and transparency. A comprehensive overview of the CSF-11-30-2A-R harmonic drive's specifications is provided in the table labeled as \ref{table:csf-11} below.

\begin{figure}[h]
  \centering
    \includegraphics[width=0.5\textwidth]{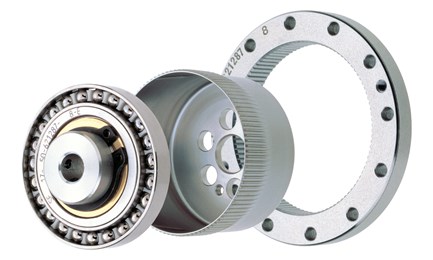}
  \caption[Harmonic Drive component set]{Harmonic Drive component set, where from left to right the components pictured are the wave generator, flex spline, and circular spline. \cite{web:harmonicDrive}}
  \label{fig:harmonicDrive}
\end{figure}

\begin{table}[h]
\vspace{5mm}
\begin{tabular}{cc}\hline
\centering
  \bf{Parameter} & \bf{Value}\\ \hline \hline
  Gear ratio & 30  \\ \hline
  Peak torque limit & 8.5 Nm \\ \hline
  Backdriving torque & 1.3 Nm \\ \hline
  Max. input speed & 8500 rpm \\ \hline
  Backlash & 2.3 $\times 10^{-5}$ rad \\ \hline
  Weight & 50 g \\ \hline
  Overall length & 25.8 mm \\ \hline
  Circular spline diameter & 40.0 mm \\ \hline
  \vspace{-5mm}
  \caption[Harmonic Drive CSF-11-30-2A-R specifications]{Harmonic Drive CSF-11-30-2A-R specifications \cite{web:harmonicDrive}}
  \label{table:csf-11}
\end{tabular}
\end{table}

A variety of DC motor types and models were considered and compared for the actuators of this robot. Options such as hydraulic and pneumatic actuators were not considered due to their weight, complexity, and need for additional supporting hardware such as pumps and pressurized tanks.  After evaluating various types of motors, a brushless DC motor stands out as a superior option. Brushless motors utilize permanent magnets on their rotor and electronically commutated coils, typically with three phases, on their stator to produce torque and motion. BLDC motor outperforms its brush counterpart and offers several key advantages over other types of motor.
The absence of brushes in BLDC motors not only reduces wear and tear but also significantly decreases maintenance requirements, leading to extended motor lifespan and enhanced system durability also it generates less loss due to minimal friction. One of the most important features of BLDC motors is their exceptional controllability due to electrical commutation which offers precise and responsive speed and torque control. Moreover, the compact, lightweight design and higher power-to-weight ratio of the BLDC motor make it ideal for our robot.

Over 40 different brushless motors were compared in order to select the most power and torque-dense motor for Harpy's leg actuators. Of all motors that were considered, the T-motor Antigravity 4006  (\cite{web:antigravity4006}) was among the most power-dense. However, it is further set apart by its torque density. Of the motors with available continuous torque measurements, the Antigravity 4006 was clearly superior in our desired continuous torque range of 300 to 500 mNm.

\section{Design overview - Aerial assembly}

\begin{figure}[h]
  \centering
    \includegraphics[width=1\textwidth]{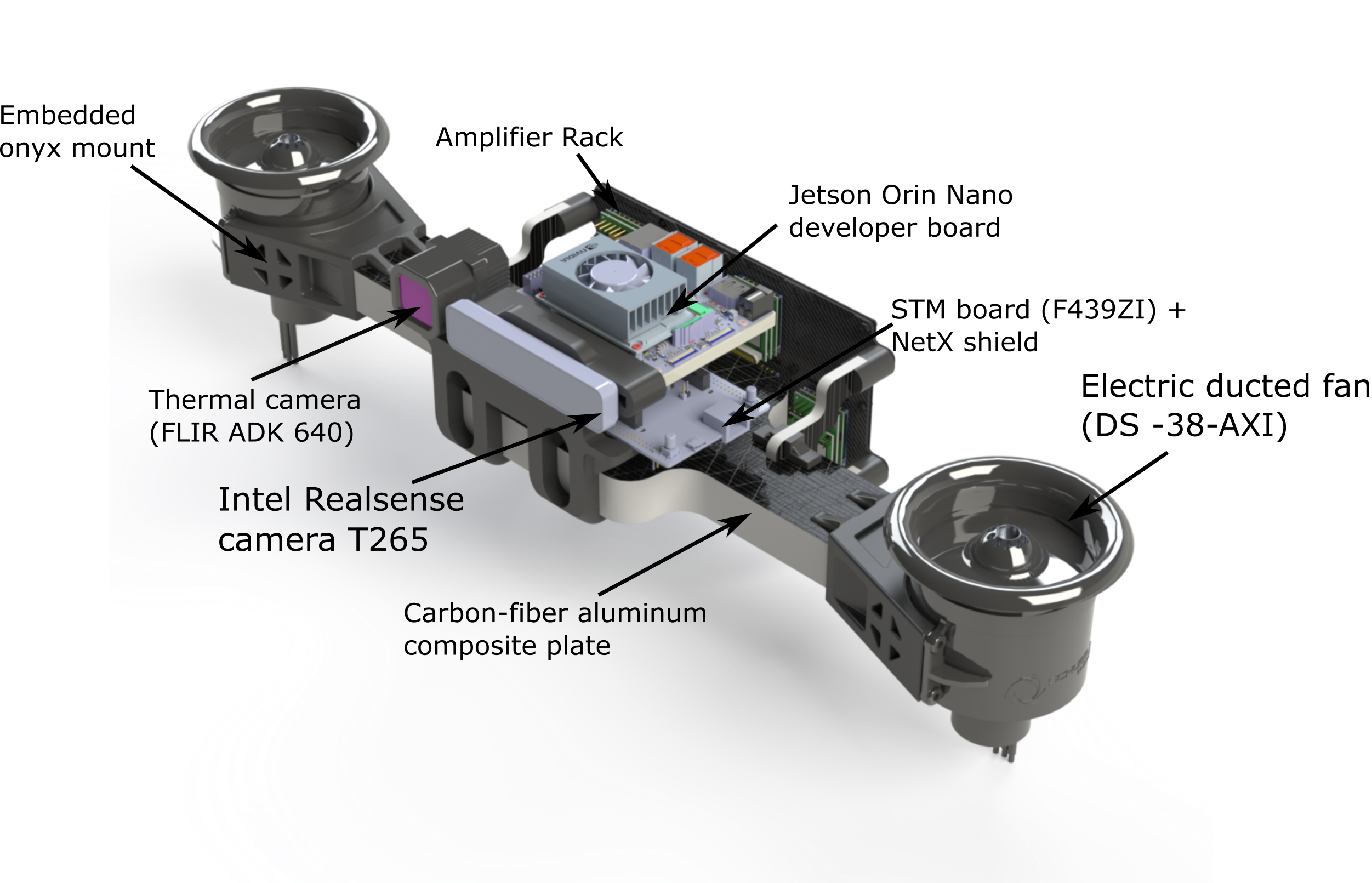}
  \caption[Harpy model]{Harpy's Thruster assembly} 
  \label{fig:Harpy upper assembly }
\end{figure}

Harpy's thrusters as shown in fig(\ref{fig:Harpy upper assembly }) are its wings, enabling it to bypass large obstacles and rough terrain in addition to giving the robot additional degrees of freedom to increase its stability. Harpy's thruster assembly consists of Carbonfiber-aluminum composite mount, a Realsense camera T265, Jetson Orin, STM(F439ZI) $+$ NetXshield board, an electric ducted fan, an amplifier rack, and IMU. This section explores the process behind selecting the type of thruster and speed controller. It will also describe the material selection for the thruster mount.

\subsection{ Mount }

The propulsion unit is a critical component for any robotic system as it enables the robot to move and traverse various terrains. It is responsible for generating the necessary force to propel the robot forward, and its design must be carefully considered to ensure maximum efficiency and stability.

The design of the propulsion unit must take into account various factors, including the weight and size of the robot, the terrain it will be operating on, and the desired speed and agility. Additionally, the propulsion unit must be strong enough to withstand continuous tests and impacts during testing. Thus, the effectiveness of the propulsion is not dependent on the type of propulsion but also depends on the mount.

For the Harpy, we designed the propulsion unit from a carbon fiber-aluminum composite plate. Before going with carbon fiber-aluminum composite, we tested all the different types and sizes of composites. We got aluminum and aramid composite of ¼ inch and ½ inch. Propulsion undergoes mostly bending deformation and thus we performed tests for bending load. 

\begin{figure}[hp]
  \centering
    \includegraphics[width=1\textwidth]{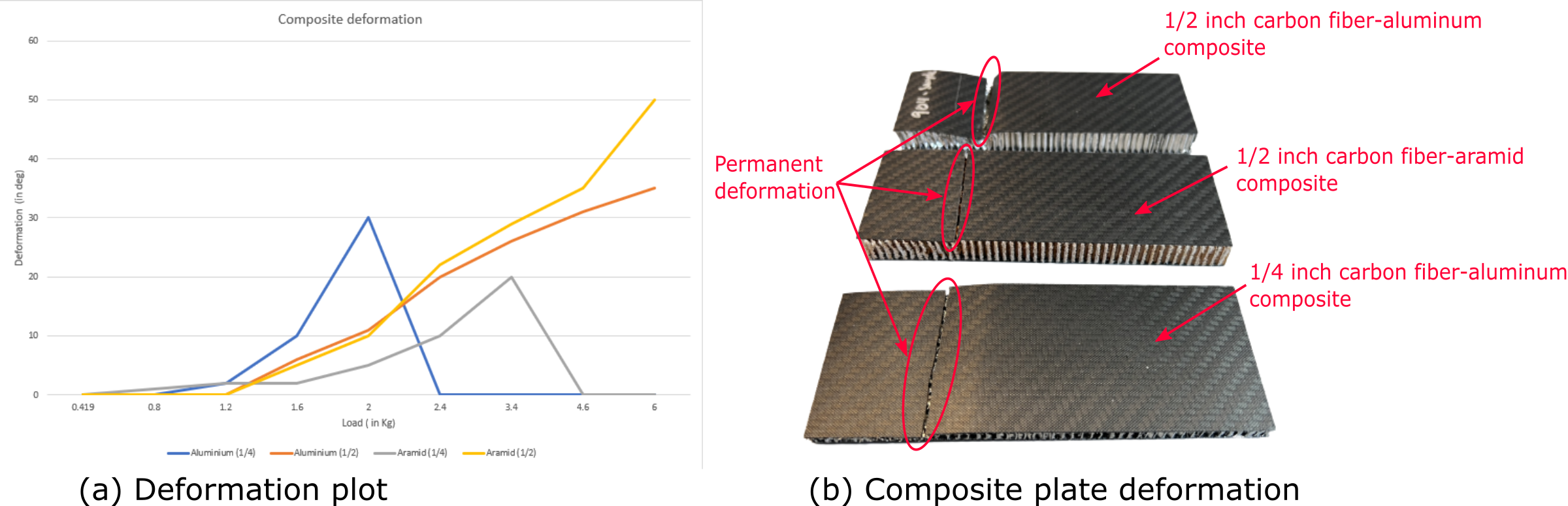}
  \caption[Composite plate deformation]{Composite plate testing} 
  \label{fig:composite plate testing}
\end{figure}

We got the sample plate of dimension $2”$ x $6”$ and thickness varied for different materials. Sample plates were fixed on one end and load was applied on the other end. We gradually applied the load and checked the amount of bending (in deg). The graph in fig(\ref{fig:composite plate testing}(a)) shows the deformation vs applied load. Both ¼-inch aluminum and $\frac{1}{4} inch$ aramid deform completely after $2 kg$ and $3.4 kg$ of static load respectively. $\frac{1}{2} inch$ aluminum and aramid composite is able to withstand about $6kg$ of load but with permanent deformation of $35 deg$ and $50 deg$ respectively.

For our application, we need composite plates to withstand about $4kg$ of thrust force which will be applied in static conditions and as an impulse force. We tested for static conditions and $\frac{1}{2} inch$ aluminum was applied to withstand it but showed significant permanent deformation. So, we decide to go with the $1 inch$ aluminum composite plate. 

\subsection{ Electric Ducted Fans}

The process of selecting the appropriate type and size of thrusters for the robot involves evaluating numerous factors including efficiency, maximum thrust, safety, response time, energy density, weight, diameter, volume, heat emission, and noise. To ensure a thrust-to-weight ratio greater than one, the target total thrust capacity for the thrusters is set at 7 kgf, equivalent to approximately 68 N.

Electric ducted fans (EDFs) stand out due to their capacity to generate significantly higher thrust outputs compared to conventional propellers of equivalent blade diameters. This feature is particularly advantageous in reducing the volume occupied by thrusters. Despite the added weight of the duct, the utilization of composite materials enables the creation of lightweight yet robust housing. Another significant advantage for ground operations, which constitute the majority of Harpy's activity, is the enhanced safety level provided by housing the blades within the duct. This design aspect reduces concerns regarding blade tip collisions, contributing to safer navigation in unknown environments and increased human safety in close proximity to the robot.

The efficiency enhancement that ducts bring to propellers is achieved by minimizing energy losses caused by blade tip vortices and converting that energy into additional thrust. The closely spaced relationship between the blade tip and the duct obstructs air passage around the blade tip due to the pressure differential between the upper and lower surfaces of the blade. Introducing a lip at the duct inlet further enhances efficiency by leveraging the Coandă effect, which generates additional thrust at no extra cost to the motor. The Coandă effect is driven by the pressure difference between the ambient air pressure and the lower pressure air moving rapidly above the curved surface. The lip's mechanism for producing thrust is analogous to the generation of lift by an airplane wing according to Bernoulli's principle, where faster-moving air over the wing's top surface creates lower pressure compared to the slower-moving air beneath.
\begin{figure}[h]
\begin{floatrow}
\ffigbox{
  \includegraphics[width=0.42\textwidth]{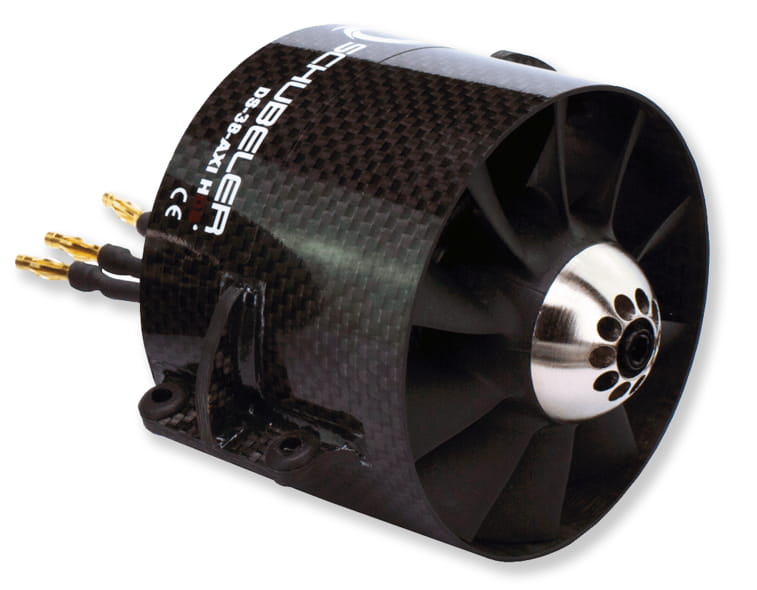}%
}{
  \caption[Schuebeler DS-38 AXI HDS]{Schuebeler DS-38 AXI HDS \cite{web:schuebelerEDF}}
  \label{fig:DS38edf}
}
\capbtabbox{%
  \begin{tabular}{cc}\hline
  \bf{Parameter(s)} & \bf{Value}\\ \hline \hline
  Max. Thrust & 53 N \\ \hline
  Max. Exhaust speed & 108.3 m/s \\ \hline
  Duct weight & 100 g \\ \hline
  Motor weight & 300 g \\ \hline
  Duct diameter & 80 mm \\ \hline
  Duct length & 60 mm \\ \hline
  \end{tabular}
}{%
  \caption[Schuebeler DS-38 AXI HDS EDF with HET 700-68-1400 motor specifications]{Schuebeler DS-38 AXI HDS EDF with HET 700-68-1400 motor specifications \cite{web:schuebelerEDF}}
  \label{table:DS38specs}
}
\end{floatrow}
\end{figure}

EDFs were chosen as the ideal thruster option, primarily due to their ability to deliver high thrust in compact volumes while also housing the blades. The Schuebeler HDS series, specifically the DS-38-AXI HDS model (Fig.~\ref{fig:DS38edf}), was selected for its remarkable energy density achieved through the combination of lightweight carbon fiber duct and blades.

For controlling the brushless DC motor connected to the EDF, an electronic speed controller (ESC) is employed. Unlike the brushless motor drives discussed in Section \ref{section:actuatorFabrication}, ESCs typically determine the rotor's position and speed by measuring back EMF during the inactive phase of the three-phase DC motor's rotation, rather than relying on position sensors like encoders or hall effect sensors. The chosen ESC for Harpy is the Castle Creations Phoenix Edge 100 ESC, known for its lightweight design (72.6 g excluding wires) and ability to deliver up to 100 A continuously at 33.6 V. Additionally, this ESC can communicate over a serial port, offering direct control of the throttle and the transmission of various data such as current, power, temperature, and speed measurements over the serial bus.

\begin{figure}[h]
\begin{floatrow}
\ffigbox{
  \includegraphics[width=0.2\textwidth]{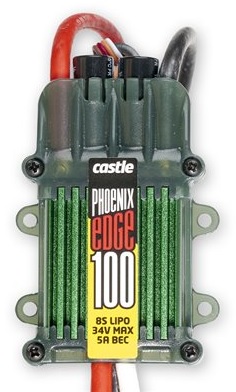}%
}{
  \caption[Castle Creations Phoenix Edge 100]{Castle Creations Phoenix Edge 100  \cite{web:ESC}}
  \label{fig:ESC}
}
\capbtabbox{%
  \begin{tabular}{cc}\hline
  \bf{Parameter(s)} & \bf{Value(s)}\\ \hline \hline
  Max. continuous current & 100 A \\ \hline
  Max. Voltage & 33.6 V \\ \hline
  Weight (w/out wires) & 72.9 g \\ \hline
  Dimensions & 51 x 72 x 23 mm \\ \hline
  \end{tabular}
  \vspace*{6mm}
}{%
  \caption[Castle Creations Phoenix Edge 100 specifications]{Castle Creations Phoenix Edge 100 specifications \cite{web:ESC}}
  \label{table:ESCspecs}
}
\end{floatrow}
\end{figure}



\chapter{Prototyping Harpy}
\label{chap:Prototyping Harpy}

\begin{figure}[H]
  \centering
    \includegraphics[width=0.8\textwidth]{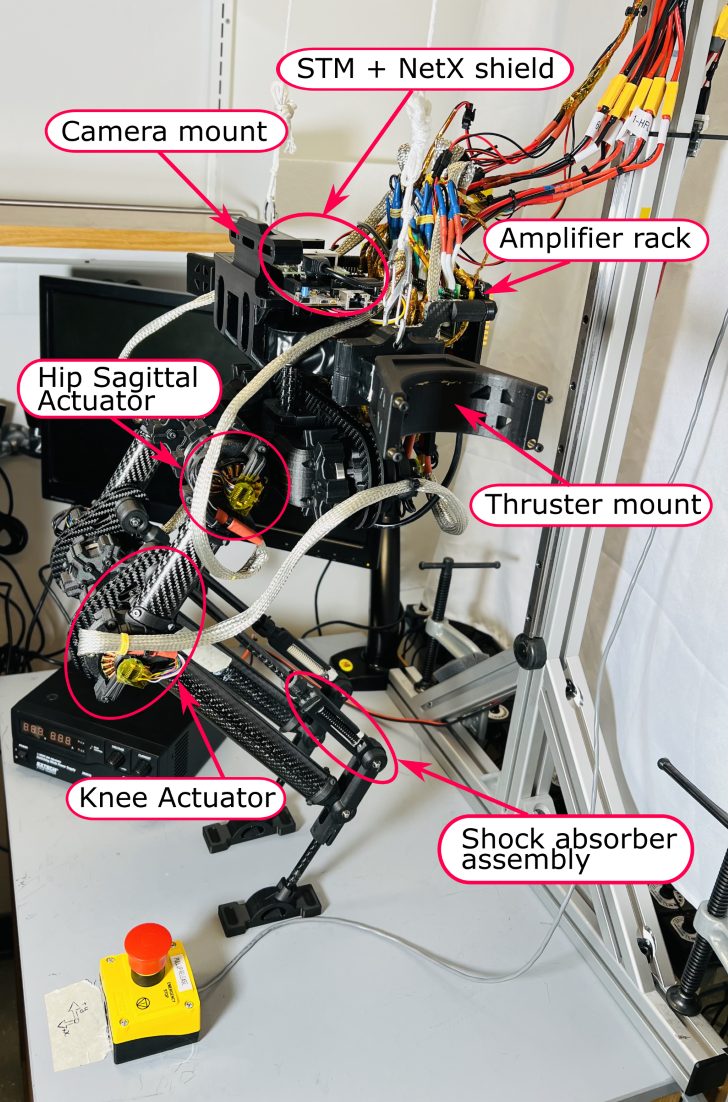}
  \caption{Harpy hardware}
  \label{Harpy hardware model}
\end{figure}

\section{3D printing}

In recent years, 3D printing technology has revolutionized the way products are designed and manufactured. With 3D printing, it's now possible to create complex shapes and designs that would have been impossible or prohibitively expensive using traditional manufacturing methods. Actuator housing undergoes a massive amount of bending moment that yields higher internal shear stress. Simply way to mitigate this stress is by applying the sandwich panel theory. It involves the use of two thin face sheets, bonded to a thicker core material, creating a sandwich structure that is much lighter and stiffer than a solid material of the same volume. This theory is particularly suitable for 3D printed materials because they often have anisotropic properties, meaning their mechanical properties can vary depending on the direction of the print layers. A simple approach to increase their stiffness while minimizing their cross-section and weight is to apply the first-order shear deformation theory. Sandwich deformation is associated with the skin’s bending and core shear. Thus, the equivalent flexural rigidity of a sandwich beam section can be calculated with the following equation.

\begin{equation}
(EI)_{eq} = \frac{E_{c}bc^3}{12} + \frac{E_{f}bt^3}{6} + \frac{E_{f}bt}{2} (c+t)^2
\end{equation}

Typically,

\begin{center}\(E_{f} \gg E_{c}\) and \(c \gg t \)
\end{center}

Here, $E_f$ is Young's modulus of skin,$E_c$ is Young's modulus of core, $t$ is the thickness of skin, $c$ is the thickness of the core and $b$ is width. This equation shows that the specific stiffness of a sandwich panel can be increased by increasing the modulus of elasticity of the face sheets and/or the core while keeping the weight low by using a lightweight core material. Overall, sandwich panel theory provides a useful framework for designing lightweight and stiff structures using 3D printed materials, and the specific stiffness equation shows how the properties of the face sheets and core can be optimized to achieve the desired performance.

\begin{figure}[h]
\begin{floatrow}
\ffigbox{%
  \includegraphics[width=0.2\textwidth]{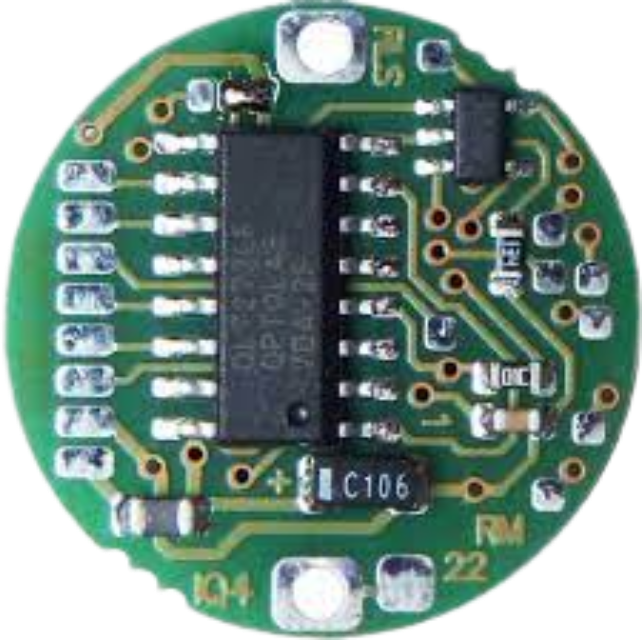}%
}{%
  \caption{RMB20 encoder module \cite{web:encoder}}
  \label{fig:RMB20}
}
\capbtabbox{%
  \begin{tabular}{cc}\hline
  \bf{Parameter} & \bf{Value}\\ \hline \hline
  V & 5V  \\ \hline
  Max. Current  & 35 mA \\ \hline
  Accuracy & 0.5 deg \\ \hline
  Resolution & 8129 \\ \hline
  \end{tabular}
\vspace*{6mm}
  }{%
  \caption[RMB 20 specification]{RMB 20 specifications\cite{web:encoder}}
  \label{table:RMB$20$IC}
}
\end{floatrow}
\end{figure}

Further, by combining 3D printing with a permanent magnet BLDC motor and harmonic drive as shown in  fig(\ref{fig:Actuator sectional viewl}), we have been able to create actuator housings that are not only customized to the specific needs of our particular application but also offer superior performance. The entire actuator assembly is further split into two subassemblies namely motor assembly and harmonic drive assembly. The motor assembly consists of a BLDC motor, encoder, top and bottom motor housing, and input shaft. Encoder specification is shown in table(\ref{table:RMB$20$IC}). The harmonic drive assembly consists of a harmonic drive, output shaft, harmonic drive housing, and flex spline clamp ring.The input shaft, output shaft, output flange, and flex spline clamp ring are machined from 316 stainless steel to ensure that they maintain an extremely precise and long-lasting outer surface so that they rotate smoothly within their bearings while transmitting high amounts of torque.

\subsection{Actuator Fabrication}
\label{section:actuatorFabrication}

\begin{figure}[h]
  \centering
    \includegraphics[width=0.65\textwidth]{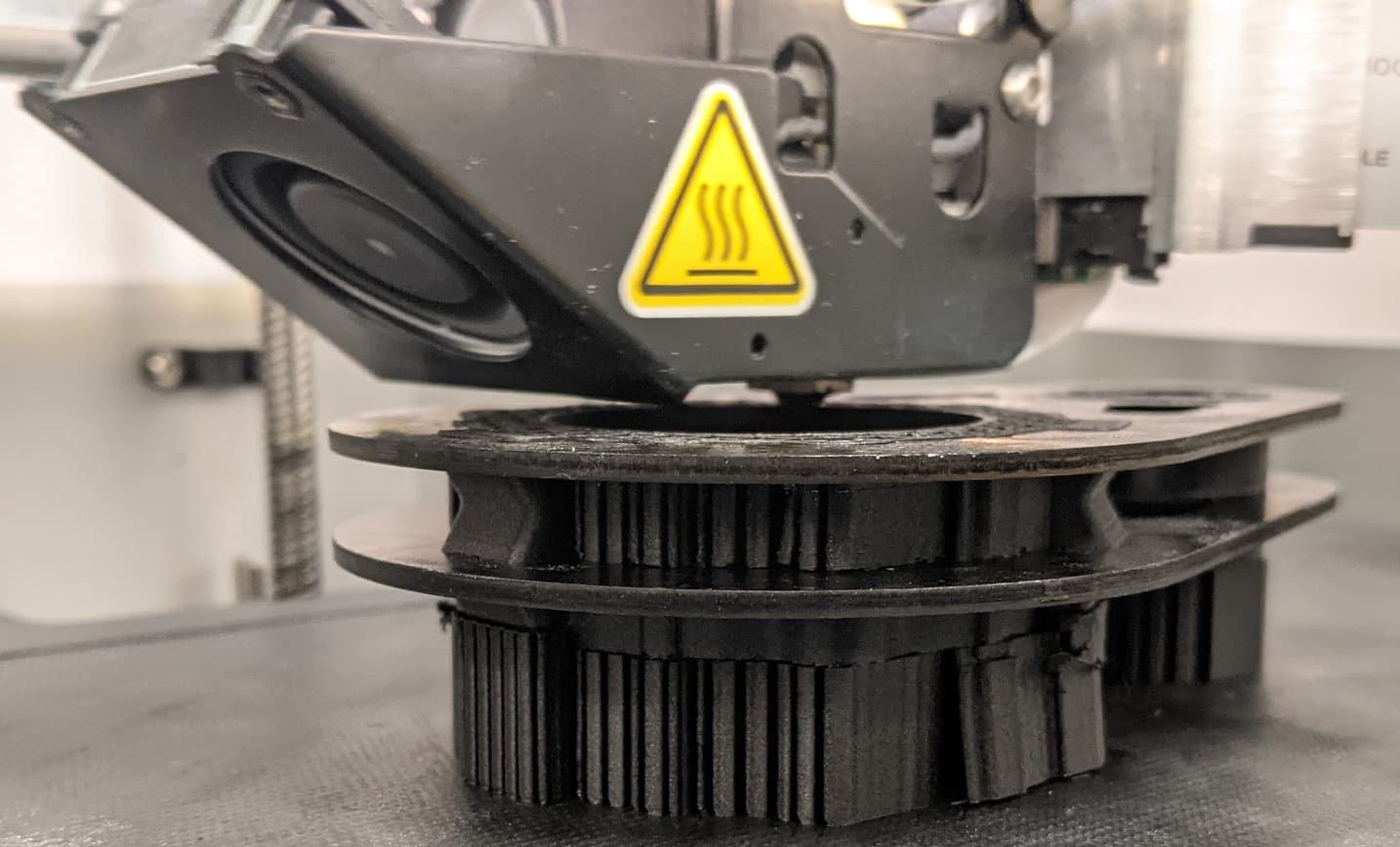}
  \caption[Markforged FDM 3D print onto an embedded carbon fiber plate]{Thermoplastic and continuous carbon fiber being deposited onto an embedded carbon fiber plate for the harmonic drive housing of the knee actuator\cite{harpy}}
  \label{fig:kneeHousingPrint}
\end{figure}

Markforged 3D printers offer a unique capability alongside their composite printing features – the ability to embed components within printed parts during the manufacturing process. This is achieved by incorporating pauses into the print, allowing the removable bed to be taken out of the printer. This strategic pause facilitates the integration of diverse components like bearings, heat-set inserts, and carbon fiber plates into the part being printed. Fig(\ref{fig:kneeHousingPrint}) exemplifies the concept with layers of thermoplastic material being deposited over a carbon fiber plate.

The layers of thermoplastic and carbon fiber not only secure these components in place but also eliminate the need for additional fasteners or retaining hardware. In comparison to alternatives such as adhesive-based methods, embedding proves to be particularly advantageous in this context. The pullout strength of the embedded part aligns with that of the surrounding material, a robust carbon fiber-reinforced plastic, rather than relying on the adhesive bond between dissimilar materials. Moreover, embedding allows for the seamless integration of components within the interior of a part, resulting in a neat and efficient design.

\begin{figure}[h]
  \centering
    \includegraphics[width=1\textwidth]{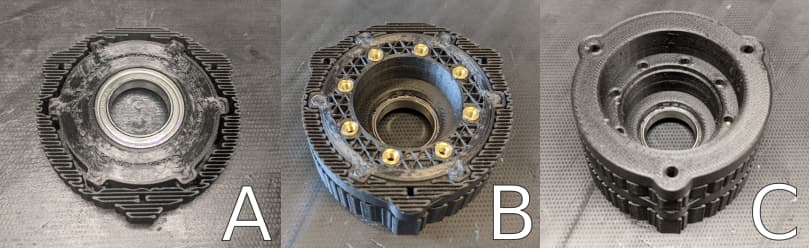}
  \caption[Leg joint harmonic drive housing embedding process]{Leg joint harmonic drive housing embedding process, where in (A) the output shaft is embedded and coated in a thin layer of glue, (B) heat-set inserts are inserted, and (C) the print is completed\cite{harpy}}
  \label{fig:legHDhousingPrint}
\end{figure}

The process of fabricating the harmonic drive housing of the leg joint actuator is illustrated in Fig(\ref{fig:legHDhousingPrint}). After the support material is removed in step A, the bearing, coated with a thin layer of polyvidone glue, is inserted in step B. This sequence is essential to prevent any contamination of the print surface, which could lead to layer delamination and compromise part strength. Following the bearing insertion, the print is finalized, and heat-set inserts are integrated using a soldering iron in the same step. Notably, the presence of additional material layers around the heat-set insert enhances its pullout strength beyond its rated capacity while simultaneously forming a flat surface for a circular spline. The final step, shown in C, showcases the completed print. It's important to mention that in cases where carbon fiber plates are embedded, as in the knee or frontal hip actuators, two additional pauses would be introduced in the print process.

\subsection{Testing Actuator}

\begin{figure}[h]
  \centering
    \includegraphics[width=1.0\textwidth]{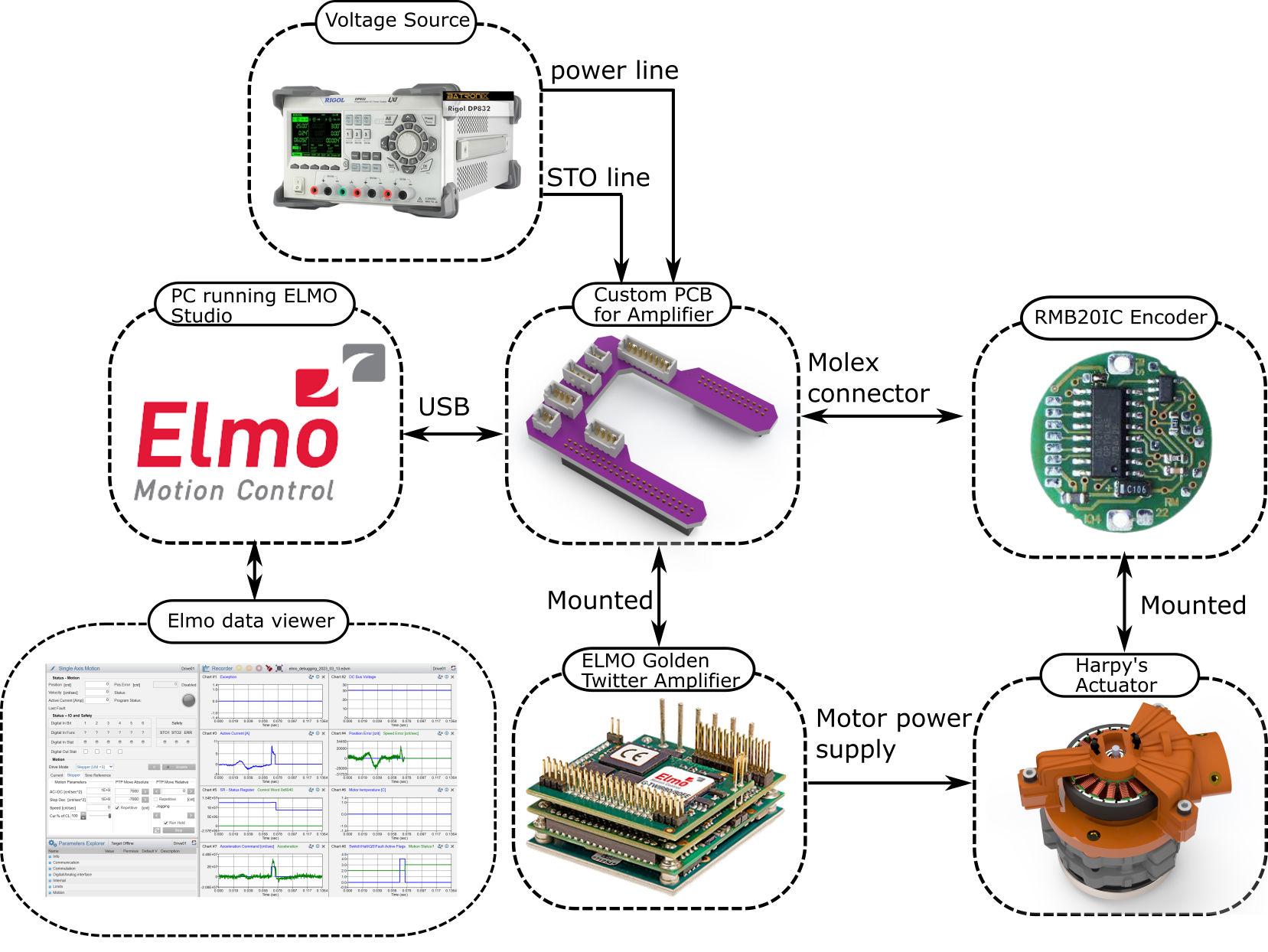}
  \caption[Harpy's actuator testing bed]{Testing flow diagram}
  \label{fig: Testing flow diagram}
\end{figure}
The joint actuators are crucial components in the bipedal system as they are responsible for the precise and controlled motion of the leg. It plays a vital role in determining the overall performance and functionality of the robot, as it directly affects the accuracy, speed, and payload capacity of the robot. To validate its functionality and performance, we employed the Elmo Studio application\cite{web:Elmostudio}, a power tool provided by the Elmo company. To test all the actuators, we created the testing bed as shown in the fig(\ref{fig: Testing flow diagram}). Testbed consists of an encoder, an Elmo amplifier with custom PCB and a voltage power source. The top side of the PCB provides EtherCAT input/output, safe torque off (STO) $5V$ signal input/output, USB input/output, and encoder input. Molex picoblade headers and wire housings are used to provide small, lightweight, and secure connections between wires and the PCB. The bottom side of the board directly interfaces with the motor drive pins using a $24$-pin and $44$-pin header.

Our testing methodology encompassed two distinct stages. The first involved the meticulous tuning of the joint actuator's parameters. This step was imperative in fine-tuning the actuator's behavior to align with our performance expectations. All of Harpy's joint actuators were tuned with the following parameters as shown in table(\ref{table:Tuning parameters}).
\newline
\capbtabbox{%
\centering
  \begin{tabular}{c c}\hline
  \bf{Parameter(s)} & \bf{Value}\\ \hline \hline
  Voltage & 30 V \\ \hline
  Peak Current & 10 A \\ \hline
  Stall Current & 8 A \\ \hline
  Max. motor speed & 8000 RPM \\ \hline
  Sensor type & Rotary \\ \hline
  Glitch factor & 10922666 \\ \hline
  Lines/revolution & 2048 \\ \hline
  Counts/revolution & 8192 \\ \hline
  \end{tabular}
}{%
  \caption[Harpy's Joint actuator tuning parameters]{ Tuning parameters for harpy's joint actuator}
  \label{table:Tuning parameters}
}

Following the parameter tuning, we proceeded to the second stage of testing. This phase involved subjecting the joint actuator to both position control and speed control assessments. The position control evaluation enabled us to ascertain the actuator's accuracy in adhering to specified positional setpoints. This assessment not only gauged the actuator's precision but also shed light on its potential for reliable and repeatable positioning tasks.

Additionally, the speed control assessment delved into the actuator's dynamic response and agility. This evaluation provided insights into its ability to attain and maintain desired speeds promptly, a crucial aspect in applications requiring rapid and controlled movement.

By undertaking these comprehensive testing procedures, we aimed to rigorously evaluate the joint actuator's performance under varying operational scenarios. The integration of Elmo Studio as the testing platform accentuated the precision and thoroughness of our assessment, thereby contributing valuable insights into the actuator's capabilities within the broader context of our research.

\section{Leg assembly}

\begin{figure}[h]
  \centering
    \includegraphics[width=1\textwidth]{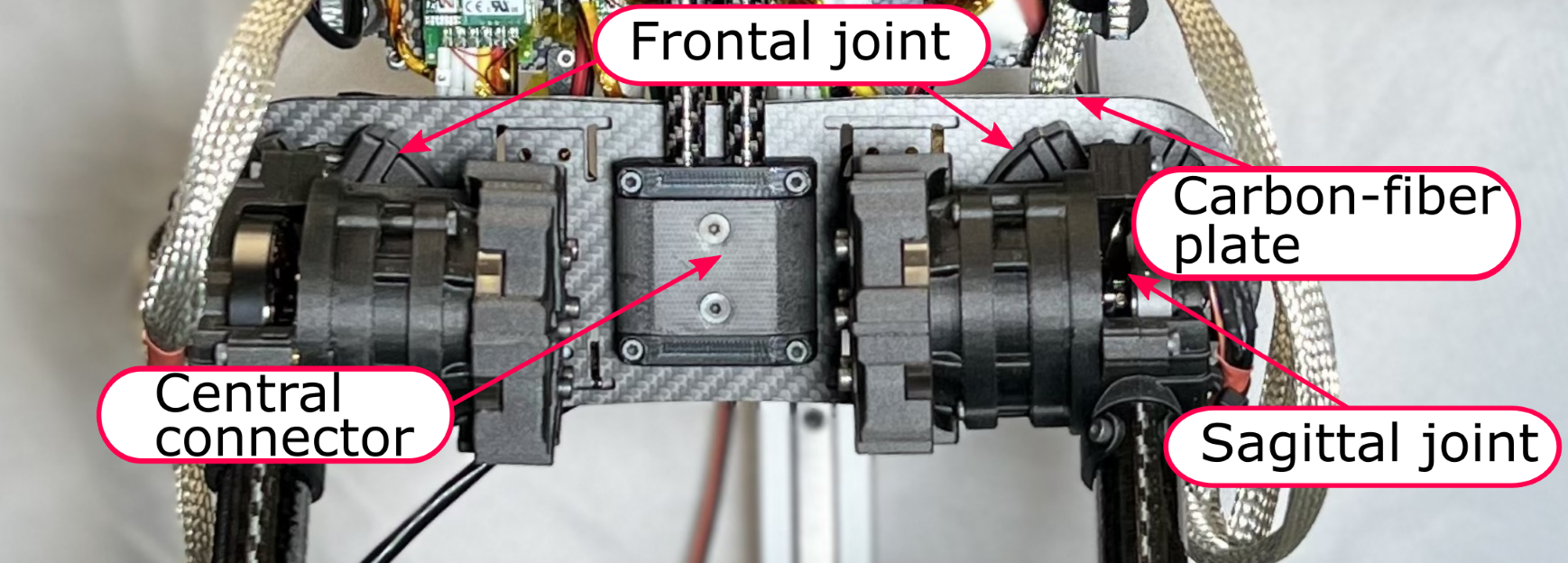}
  \caption{ Harpy's pelvis block}
  \label{fig:pelvis block}
\end{figure}

After the successful testing and manufacturing of the actuator, the subsequent phase involved the production of other integral components within the leg assembly, such as the Pelvis block and shock absorber assembly. As illustrated in Fig (\ref{fig:pelvis block}), this assembly encompasses a pair of frontal joints, which were meticulously fabricated and subjected to testing, as expounded upon in Section \ref{section:actuatorFabrication}.

Additionally, a central connector was 3D printed to secure the mounting of the thruster assembly. Both the frontal joints and the central connector were strategically integrated into two carbon fiber plates, each measuring 2 mm in thickness. These carbon fiber plates not only provided structural support but also contributed to the overall robustness of the assembly.

\begin{figure}[h]
  \centering
    \includegraphics[width=0.7\textwidth]{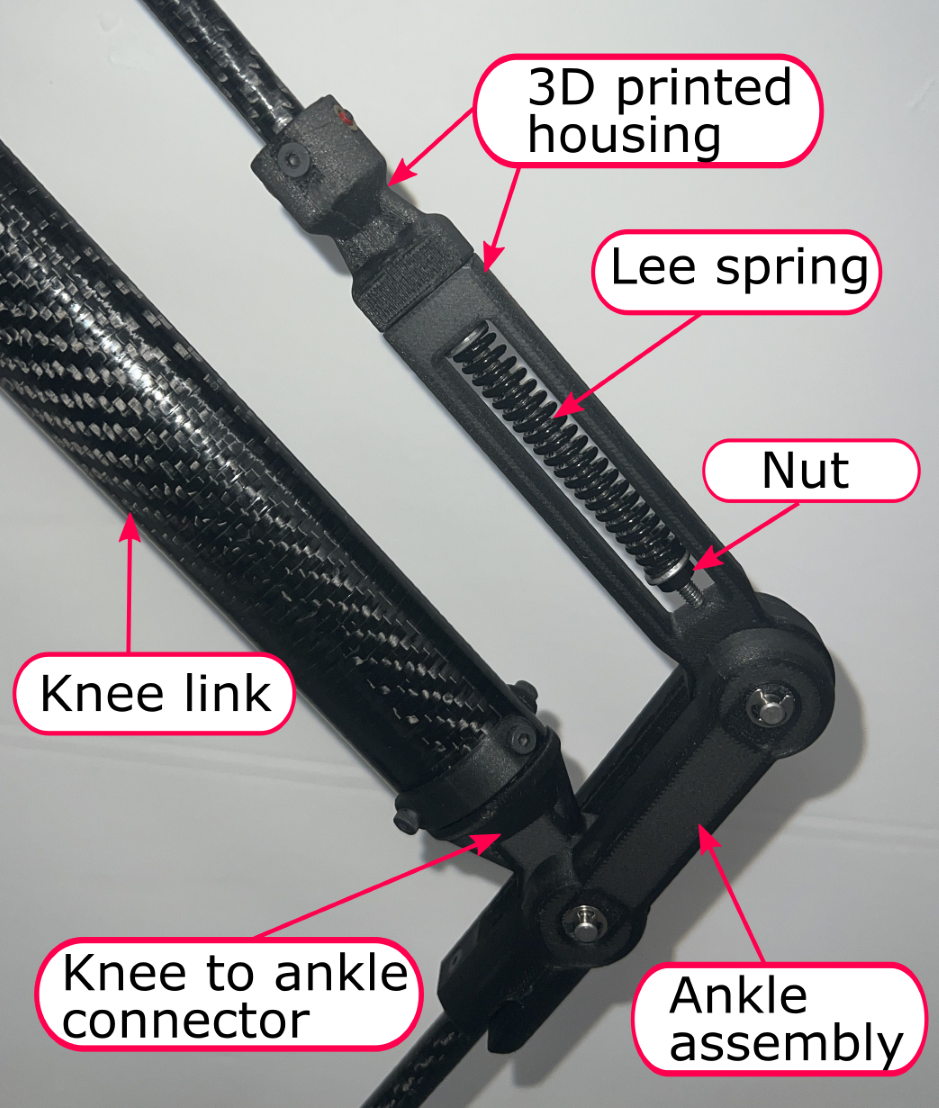}
  \caption{ Harpy's ankle joint}
  \label{fig:Ankle joint}
\end{figure}

Subsequent attention was devoted to the construction of the shock absorber assembly as shown in fig(\ref{fig:Ankle joint}). This assembly comprised several key components, including a lead screw, nut, spring(LHL 375AB), and a specially designed 3D-printed housing. The housing was manufactured using a 3D printing process, with the layers oriented perpendicular to the anticipated direction of force application. To enhance its resilience and capacity to absorb shocks, the housing was further fortified with Kevlar, a high-strength synthetic fiber renowned for its exceptional durability and impact resistance.

By meticulously applying a controlled preload to the system, the harpy robot was endowed with the ability to traverse its environment in a manner reminiscent of rigid-legged locomotion. Drawing from past successes in the realm of legged robotics, the incorporation of springs in series with a pantograph leg has proven effective in mitigating the peak torques experienced by actuators during drop tests of legged robots \cite{conference:springQuadruped}\cite{paper:seriesElasticLeg}.

In summary, the evolution of the leg assembly from the actuator phase to the inclusion of the Pelvis block and shock absorber assembly showcases a meticulous progression towards enhancing the robustness, efficiency, and overall performance of the harpy robot's locomotion capabilities.

\section{Thruster Mount}

\begin{figure}[h]
  \centering
    \includegraphics[width=1\textwidth]{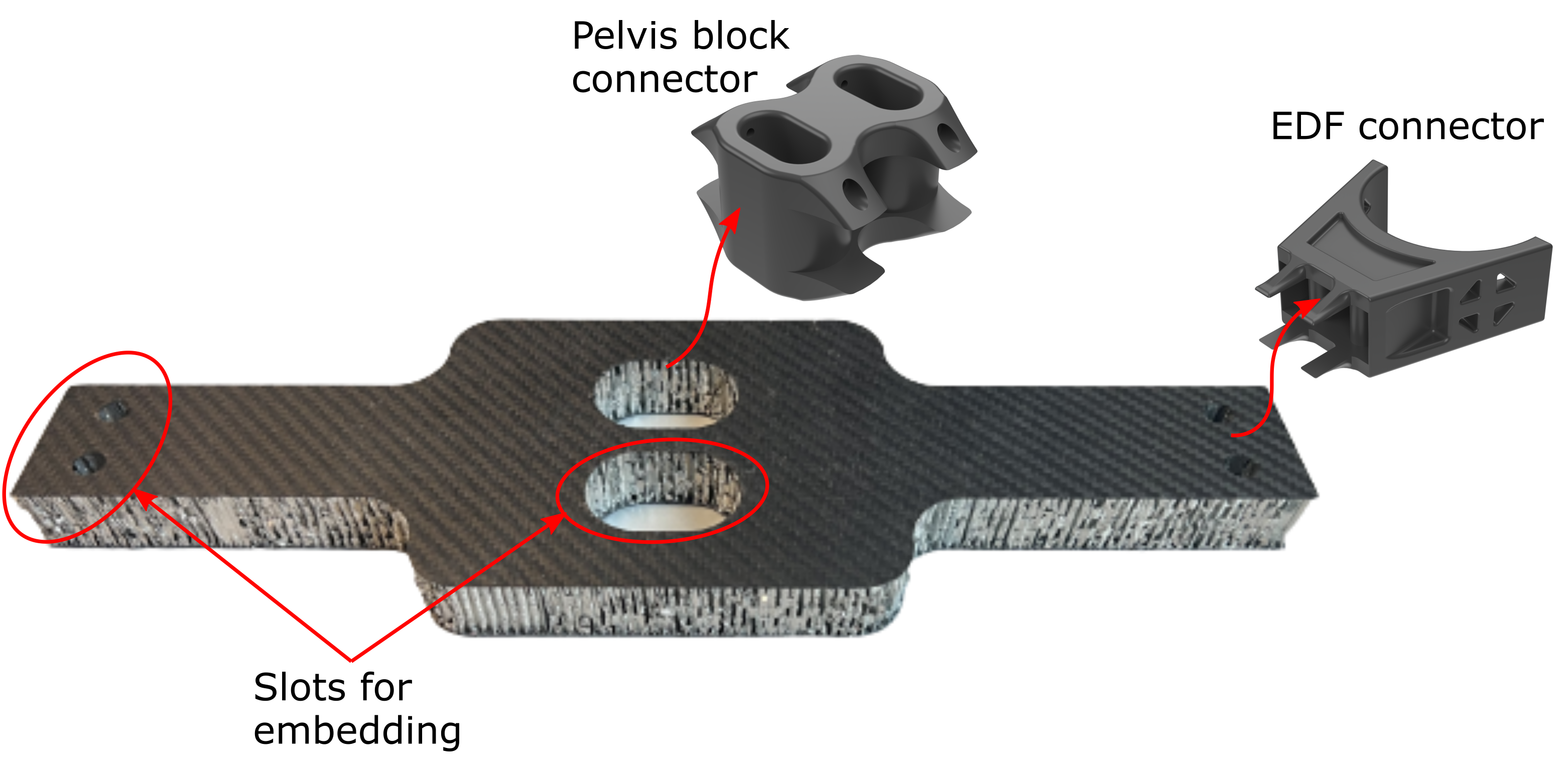}
  \caption{ Carbonfiber-Aluminum Thruster Mount}
  \label{fig:thruster mount}
\end{figure}

In Chapter(\ref{chap: Northeastern's Harpy platform}), we delve into the intricacies of Northeastern's Harpy platform and its innovative thruster mount design. This mount stands out due to its incorporation of composite materials, a choice motivated by its numerous benefits. Our primary objective was to ensure a lightweight mount and to achieve this, we strategically embedded connectors for EDFs (Electric Ducted Fans) and a pelvis block within it.

For the manufacturing process, we turned to the capabilities of a Markforged 3D printer, due to its unique ability to pause the print and embed components during the fabrication process and reinforce 3D printed parts. This attribute was pivotal in realizing our design aspirations. The resulting thruster mount, a pivotal component of the Harpy platform, is vividly illustrated in Figure(\ref{fig:thruster mount}.

However, the size of the thruster mount posed a challenge, as attempting to print it in its entirety in one go was not feasible. To address this, we adopted a phased printing approach, depicted step by step in Figure(\ref{fig:thruster mount}). The sequential printing approach commenced with the fabrication of the central connector, a foundational element of the mount's architecture.

Recognizing the critical demands placed on 3D printed components by the operational dynamics of a thruster, we didn't stop at the basic fabrication. To fortify the 3D-printed parts, we integrated Kevlar reinforcement. This strategic addition aimed to bolster the structural integrity of the mount, providing resilience against the impulse loads generated by the thruster's operation.

\section{Electrical Communication between modules}

The firmware for Harpy is developed using Simulink and is then deployed to a designated PC running Simulink Realtime. This setup enables real-time execution of the control algorithms. Modifiable block parameters within MATLAB can be adjusted while the target system is operational, allowing for dynamic control over the trajectories and commands transmitted to the robot. In terms of hardware, the Nucleo microcontroller is interconnected in a daisy chain configuration with the Elmo Gold Twitter motor drivers. This arrangement establishes a seamless communication pathway between the microcontroller and the motor drivers, facilitating coordinated control and operation.

\begin{figure}[h]
  \centering
    \includegraphics[width=1.0\textwidth]{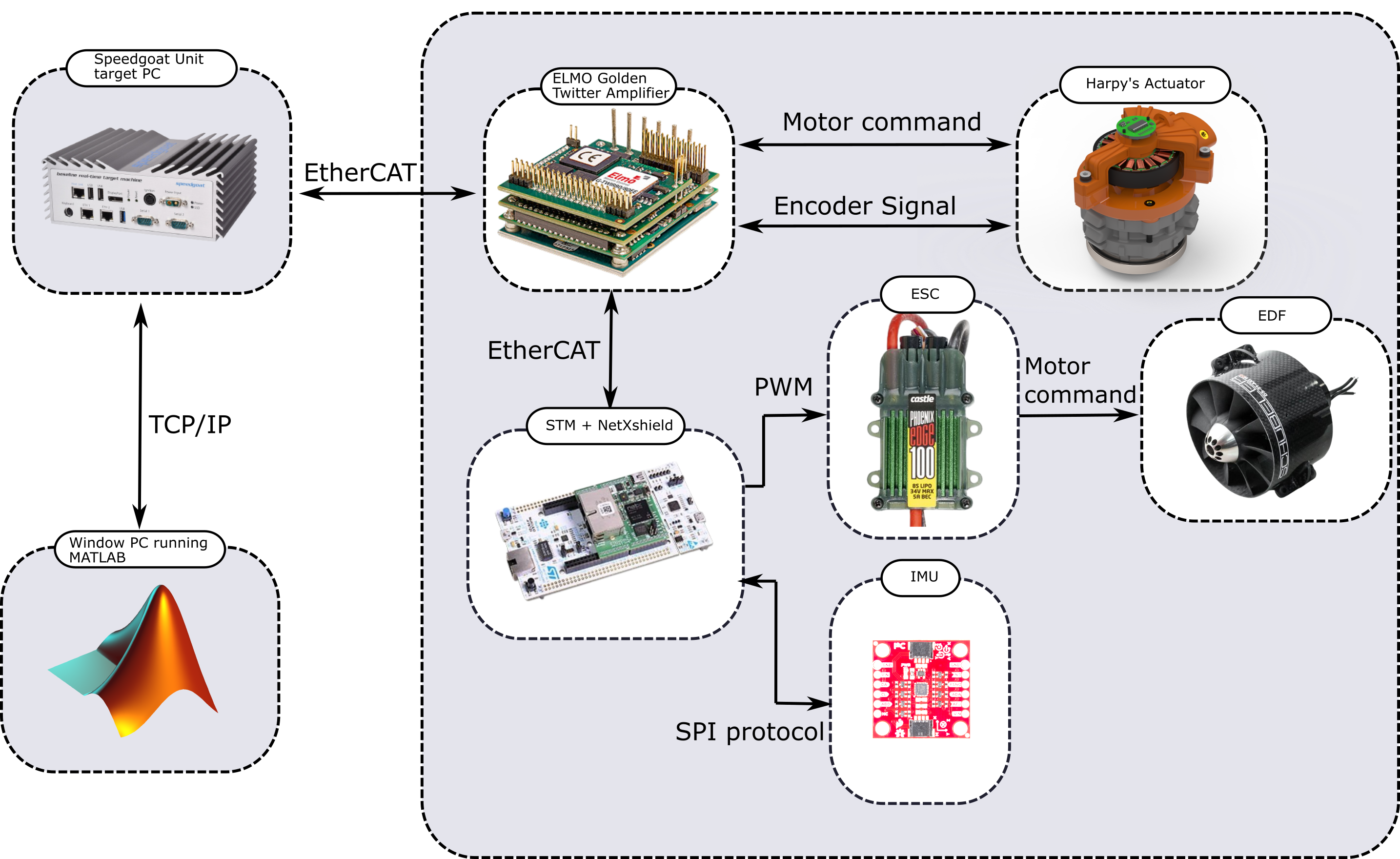}
  \caption[Robot communication signal diagram]{Robot communication signal diagram}
  \label{fig:communicationArchitecture}
\end{figure}

\begin{figure}[h]
  \centering
    \includegraphics[width=1.0\textwidth]{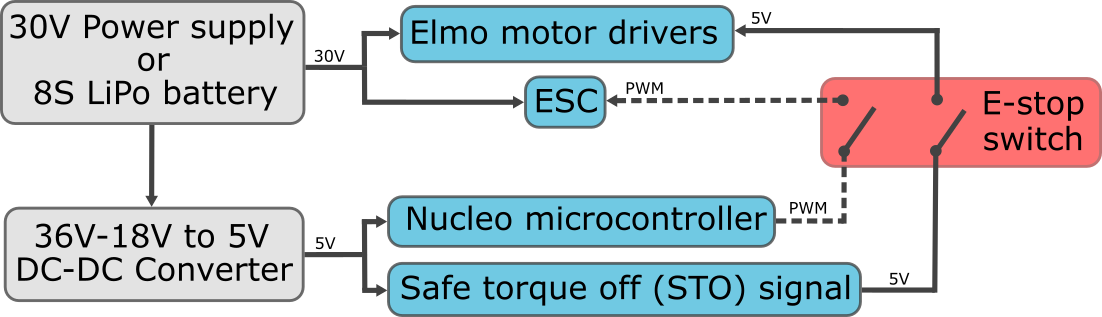}
  \caption[Power supply and emergency system diagram]{Power supply and emergency stop system diagram}
  \label{fig:powerArchitecture}
\end{figure}

Figure (\ref{fig:communicationArchitecture}) depicts the main sub-modules and their communication. The system comprises two major controllers: a low-level and a high-level controller. The low-level controller runs at a rate of 500Hz on a real-time processor from Speedgoat, a custom computer with serial and EtherCAT communication capabilities through four RS232 ports and an EtherCAT compatible chipset. A Sparkfun ICM20948 provides orientation feedback to the low-level controller via the Nucleo microcontroller and NetX shield. The high-level controller runs ROS (Robot Operating System) on an NVidia Jetson Nano and uses an Intel Realsense stereo-camera and IR camera along with IMU data for path planning and navigation. In addition, the Nucleo microcontroller also interacts with ESC to control EDF. The target unit communicates with the motor drives via EtherCAT to ensure fast update times and precise synchronization. The Master (Speedgoat) issues position commands to 6 ELMO motor drives at a rate of 10kHz. The motor driver converts the input signal into PWM signals, which control the current transmitted to each of the three phases of the brushless motors. The drives also output signals from magnetic incremental encoders at a rate of 500-4kHz (depending on the maneuvers considered). fig(\ref{fig:powerArchitecture}).

\section{Inverse kinematics}
\label{Inverse kinematics}

\begin{figure}[h]
  \centering
    \includegraphics[width=1.0\textwidth]{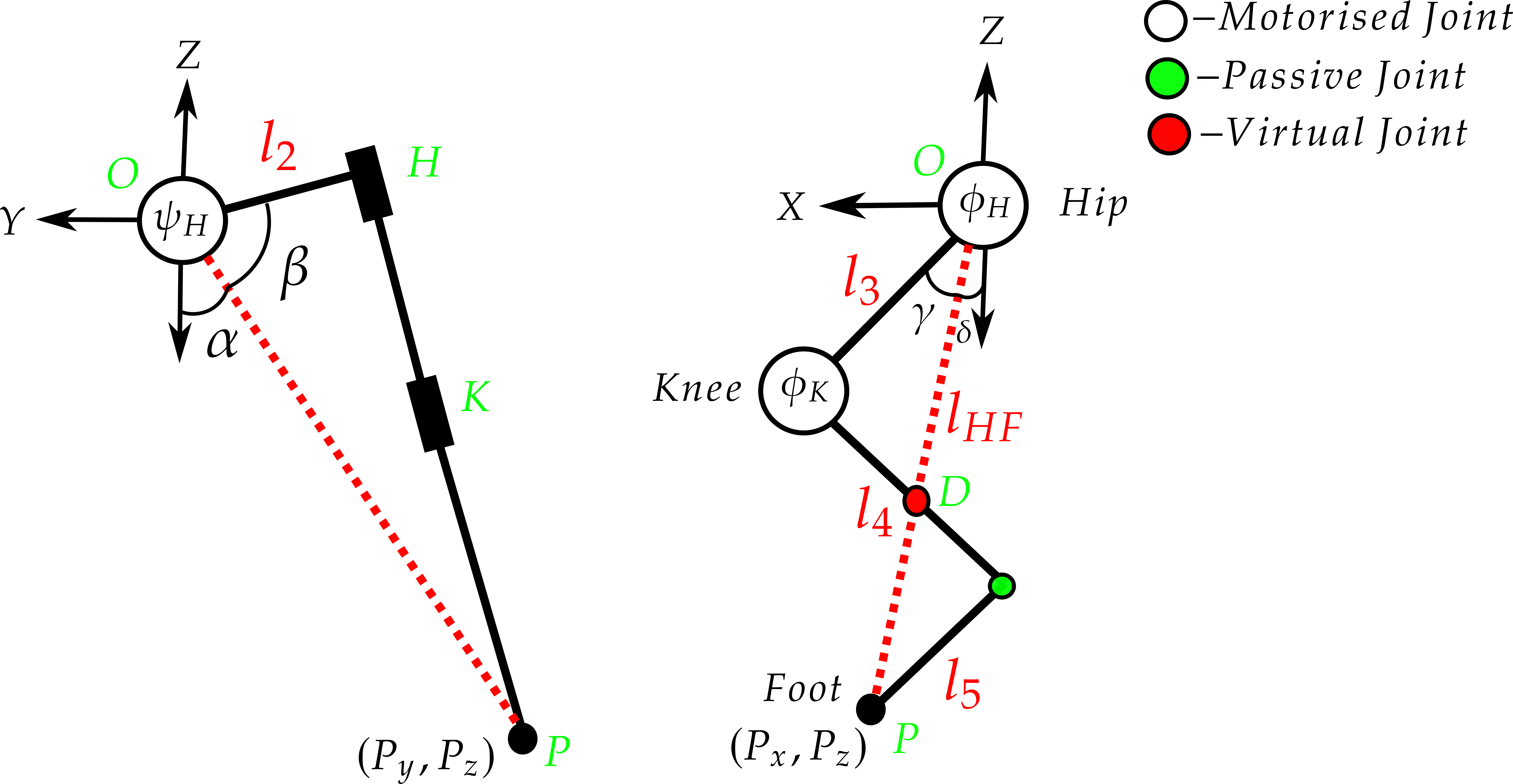}
  \caption[Harpy's HROM model]{Reduced Order Model for Harpy}
  \label{fig:HROM}
\end{figure}

Bipedal robots rely on the concept of inverse kinematics to achieve precise control and coordination. Inverse kinematics enables the determination of joint angles necessary for positioning the robot's end effectors, like its feet, in desired locations and orientations. Unlike forward kinematics, which calculates end-effector positions from joint angles, inverse kinematics addresses how to manipulate the robot's joints to achieve specific tasks, making it fundamental for tasks such as walking, running, and interacting in dynamic environments. Harpy reduced order model is shown in fig(\ref{fig:HROM}). There are three motorized joints namely hip frontal joint ($\psi_H$), Hip sagittal joint ($\phi_H$), and Knee joint ($\phi_K$). For precise feet positioning, we need to find these three joint angles.

To obtain the hip frontal angle($\psi_H$), we need to first find angle made by virtual line OP with respect to Z-axis.
\begin{equation}
\alpha = tan^{-1}(\frac{P_y}{P_z}),
\end{equation}
Next, $\triangle OHP$ is a right angled triangle and thus we can find angle $\beta$ using cosine rule.
\begin{equation}
\beta = Cos^{-1}(\frac{l_2}{\sqrt{(P_z)^2+(P_y)^2}}),
\end{equation}
Lastly, Hip frontal angle ($\psi_H$) is summation of equation (2) and equation (3).
\begin{equation}
\psi_{H} = \alpha + \beta,
\end{equation}
Equation(4) is hip frontal angle for left leg and for right leg, it will be, 
\begin{equation}
\psi_{H} = \pi - \alpha + \beta,
\end{equation}
As discussed in chapter(\ref{chap: Northeastern's Harpy platform}), Harpy's leg is designed to have a pantograph structure. This particular configuration makes the inverse kinematics calculation simplier. When we create a virtual line connecting point $O$ and $P$, we get two similar triangles namely, $\triangle ODK$ and $\triangle PDA$. Further, we can use property of similar triangle which states that two similar triangle have their length proportional and angles congruent. Thus, we get below equation using similar triangle property,

\begin{equation}
\frac{l_3}{l_5}=\frac{l_{HD}}{l_{DP}}=\frac{l_{KD}}{l_{DA}}
\end{equation}
Since we know length $l_{3}$ and $l_{5}$, we can find $l_{HD}$,$l_{DP}$,$l_{KD}$ and $l_{DA}$. Thus we know all the lengths of two triangle and can easily find angles. Knee Angle ($\phi_K$) can be found out by using cosine rule.
\begin{equation}
\phi_k = Cos^{-1}(\frac{(l_3)^2+(l_{KD})^2-(l_{HD})^2}{2l_{3}l_{KD}})
\end{equation}
Angle $\gamma$ and $\delta$ collective creates Hip sagittal angle($\phi_H$)
\begin{equation}
\gamma = Cos^{-1}(\frac{(l_3)^2+(l_{HD})^2-(l_{KD})^2}{2l_{3}l_{HD}}
\end{equation}

\begin{equation}
\delta = tan^{-1}(\frac{P_x}{P_z})
\end{equation}

\begin{equation}
\phi_{h} = \gamma + \delta 
\end{equation}

Calculations for Hip sagittal($\phi_H$) and Knee angle($\phi_K$) are similar for both right and left leg. Using equation (4), equation (6) and equation(9) we created a matlab script which we further use on simulink real-time model and in Simscape model.


\chapter{Controller Design}
\label{chap: Controller design}

\section{Control challenges}

The control of bipedal robots poses a multitude of challenges that must be addressed to achieve stable and efficient locomotion. In this section, we delve into the control challenges inherent to the bipedal robotic system, highlighting the complexities arising from the need to emulate the human-like motion while contending with inherent mechanical constraints and environmental uncertainties.
\begin{itemize}
    \item Dynamic stability: Dynamic stability is a fundamental concern in the field of bipedal robotics, as it underpins the ability of two-legged robots to maintain upright posture and navigate complex environments. Due to the inherent instability of the two-legged configuration, controlling the center of mass(COM) trajectory becomes crucial. Real-time adjustments\cite{footplacementstrategy} are required to prevent toppling and to counteract disturbances introduced by sudden external force and uneven terrain. Ensuring that the robot responds to this perturbation promptly while still maintaining its equilibrium becomes a fundamental control challenge.
    
    \item Sensor integration and feedback: Bipedal robots rely on sensor feedback to perceive their surroundings and adjust their movements accordingly. Accurate and timely sensor data, such as inertial measurement units (IMUs) and proprioceptive sensors, is essential for effective control. Developing robust control algorithms that seamlessly integrate sensory information to regulate joint angles, torques, and foot placements is crucial for achieving reliable and robust locomotion.
    
    \item Human-like Locomotion: Achieving human-like locomotion is a long-standing aspiration in bipedal robotics. This entails precise control of joint angles, coordination of limbs, and the replication of natural walking and running patterns. Developing control strategies that generate motion resembling human locomotion while accounting for biomechanical differences is a formidable challenge requiring synergy of robotics, biomechanics, and control theory.
\end{itemize}

\section{Notion of stability}
\begin{figure}[h]
  \centering
    \includegraphics[width=1.0\textwidth]{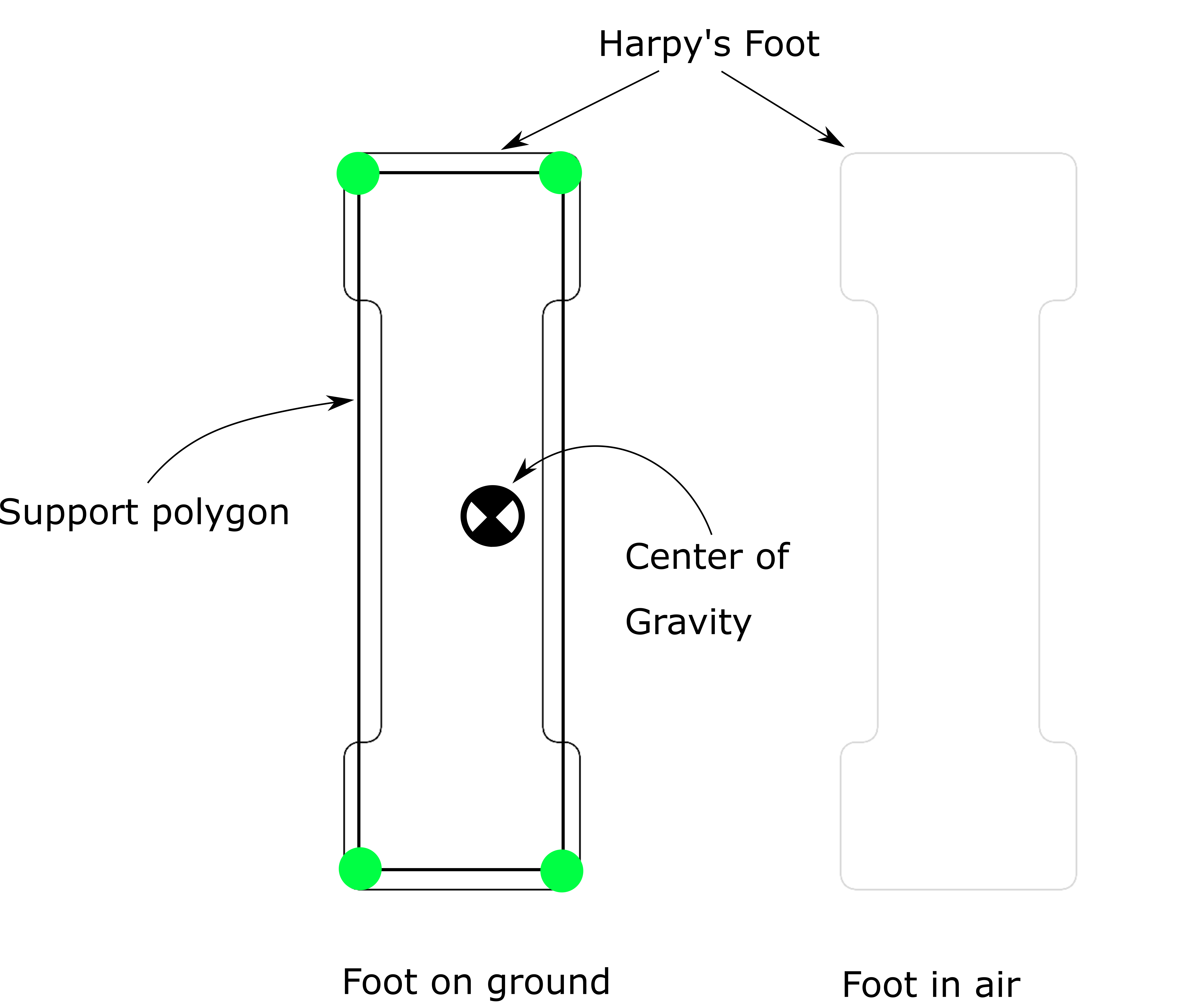}
  \caption[Harpy's support polygon]{Harpy's support polygon}
  \label{fig:support polygon}
\end{figure}

The concept of stability in bipedal robots is of paramount importance as it underpins their fundamental ability to maintain a balanced and upright posture, both during stationary poses and dynamic locomotion. Unlike stable structures or quadrupedal systems, bipedal robots inherently exist in a state of instability due to their two-legged structure. Stability in bipedal robots encompasses two key dimensions: static stability and dynamic stability. Static stability refers to the ability of the robot to maintain its equilibrium while at rest, ensuring that its center of mass remains within a stable range over its support polygon. Dynamic stability, on the other hand, pertains to the robot's capability to manage the dynamic shifts in its center of mass as shown in fig(\ref{fig:support polygon}) that occur during movements such as walking, running, or changing directions.

Dynamic stability in bipedal locomotion is a critical consideration, necessitating precise control strategies. One prominent method for maintaining stability involves manipulating the Zero Moment Point (ZMP). The ZMP, a pivotal reference point on the ground, ensures that the net moment around any axis is balanced, contributing to overall stability. To ensure dynamic stability, the ZMP must lie within the support polygon, defined by the points of contact between the robot's feet and the ground \cite{conference:ZMP}.

ZMP control is a strategic approach that entails adjusting the positions of foot placements and the distribution of ground reaction forces beneath the feet. The objective is to ensure that the ZMP remains within a predetermined stable region. This strategy aids the robot in sustaining balance and prevents tipping during walking or other locomotion modes.

In parallel, the capture point theory offers an effective avenue in the realm of bipedal locomotion \cite{web:Capturepoint}. Initially introduced by Shuuji Kajita, this theory utilizes a mathematical framework grounded in the 3D Linear Inverted Pendulum (LIP) model to generate walking patterns. The capture point signifies a specific point in 3D space, predicting the ideal position for the robot's Center of Mass (COM) to maintain stability.

Building upon Kajita's work, Johannes Englsberger developed a controller rooted in capture point dynamics (\cite{web:ControllerbasedonCP}). This controller harnesses the insights from capture point theory and demonstrates the exponential stability of its performance.

Jerry Pratt extended the notion of the capture point beyond a single ground point to encompass an entire stable region (\cite{web:Humanoidpushrecovery}). Pratt accomplished this by incorporating a flywheel into the LIP model. Leveraging the flywheel's angular momentum, Pratt effectively enhanced the robot's stability during movement.In summary, the management of dynamic stability in bipedal locomotion involves two central strategies: ZMP control for adjusting footstep positions and ground reaction forces, and capture point theory for determining optimal COM positions. Pioneered by researchers such as Shuuji Kajita, Johannes Englsberger, and Jerry Pratt, these concepts contribute significantly to advancing stable and efficient bipedal robot locomotion.

\section{Capture point theory controller}

\begin{figure}[h]
  \centering
    \includegraphics[width=1.0\textwidth]{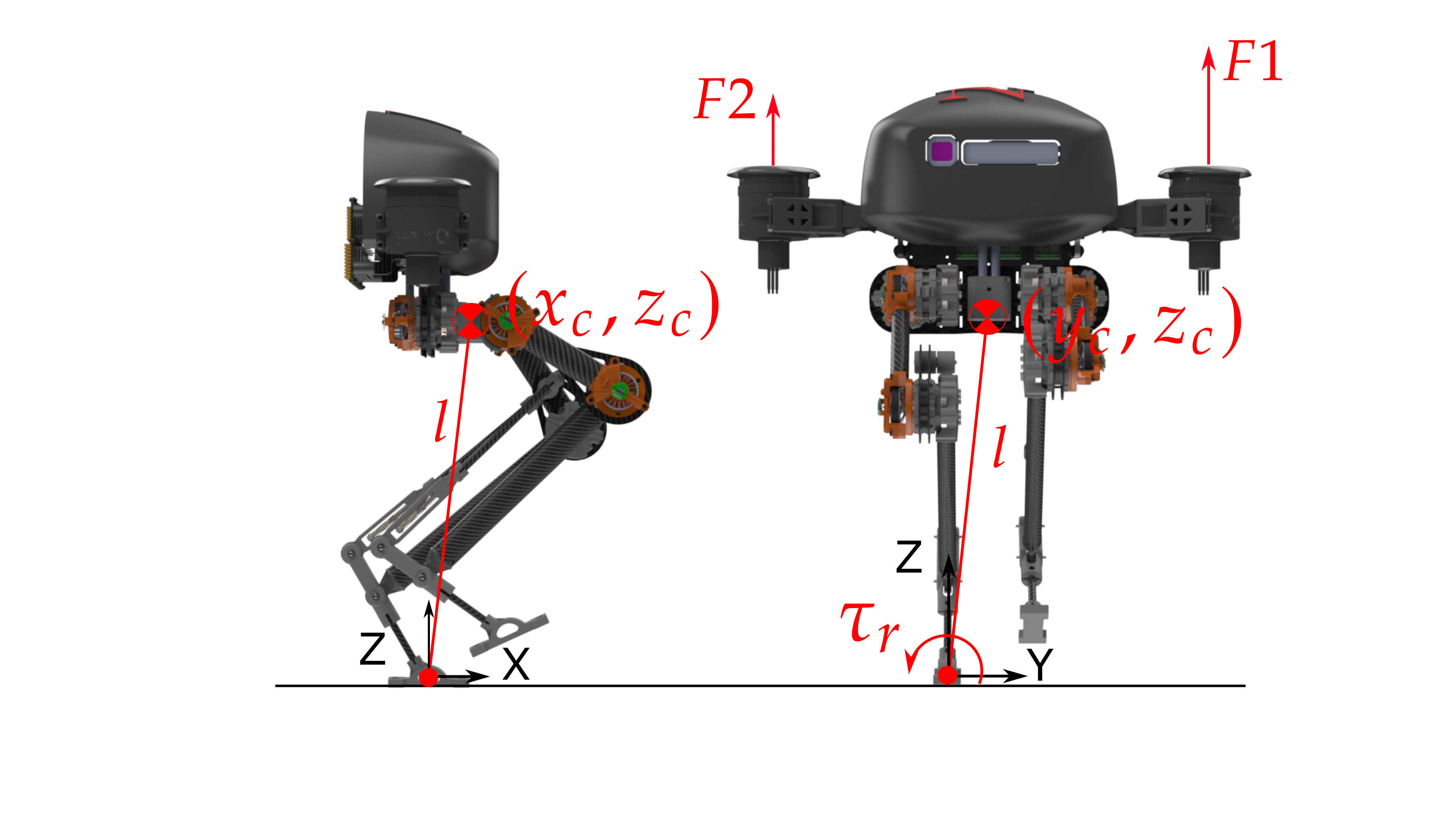}
  \caption[3D linear inverted pendulum]{Harpy modeled as 3D linear inverted pendulum}
  \label{fig:Capture point}
\end{figure}

During the swing phase, bipedal dynamics can be summarized as a Linear inverted pendulum model. We begin with our bipedal system which is modeled as a $3$D linear inverted pendulum with thrusters as shown in fig(\ref{fig:Capture point}). $f1$ and $f2$ are the thrust force generated by the thrusters. The thruster force produces a net moment about the y-axis. The model has mass $m$, length $l$, and a center of mass located at a height $z_c$ from the pivot point. The pitch angle is $\theta_p$ and the roll angle is $\theta_r$. The equation of motion during the single support phase can be written as

\begin{equation}
    \ddot{x_c} = \frac{g}{z_c}(x_c),
\end{equation}
\begin{equation}
    \ddot{y_c} = \frac{g}{z_c} (y_c) - \frac{\tau_r}{mz_c},
\end{equation}

where $(x_c, y_c, z_c)$ are the capture point coordinates,$\tau_r$ is the net torque about x-axis and $g$ is the acceleration due to gravity.The above equations are all linear given the $z$ is constant. This linearity makes the linear inverted pendulum with thruster valuable for analysis.

\subsection{Derivation of Capture Point for Linear Inverted Pendulum without thrusters }

First, we compute the equation of motion for a Linear inverted pendulum. This can be done by substituting $\tau_r =0 $ in equation (2).

\begin{equation}
    \ddot{x_c} = \frac{g}{z_c}(x_c),
\end{equation}
\begin{equation}
    \ddot{y_c} = \frac{g}{z_c} (y_c),
\end{equation}

Orbital energy \cite{conference:orbitalEnergy} is a conserved quantity that characterizes the energy of a bipedal robot's locomotion as it moves along its walking trajectory. The orbital energy is conserved during the robot's motion because the sum of its potential and kinetic energies remains constant. Orbital energy is governed by the kinetic energy (KE) and potential energy (PE) of the system.

\begin{equation}
    E = \frac{1}{2}m(\dot{x})^2 - \frac{1}{2}\frac{mg}{z_c}(x_c)^2,
\end{equation}

In equation(5), the first term represents the kinetic energy and the second term represents the potential energy. As mentioned in \cite{web:Humanoidpushrecovery},If E $>$ $0$, it means that COM is moving toward the foot and the robot can stabilize itself by taking additional steps. If E $<$ $0$, it means COM is moving away from the foot and in the opposite direction. The robot is unstable. When E $= 0$, the robot will come to rest over the foot. Substituting E = $0$ in equation (5) provides us capture point coordinate for x-direction.
\begin{equation}
    x_{capture} = \dot{x}\sqrt{\frac{z_c}{g}},
\end{equation}
Since the model is linear all the equations derived for the X-direction, can also be applied for the Y-direction. The capture point in Y-direction will be,
\begin{equation}
    y_{capture} = \dot{y}\sqrt{\frac{z_c}{g}},
\end{equation}
 Currently, Harpy doesn't have thruster on board and thus, we ignore the torque term for the Harpy controller and use the traditional capture point equation.

\section{Bezier curve gait design}
Bezier curves serve as a fundamental tool to design paths that seamlessly transition between points, ensuring continuity and smoothness. By adjusting the positions of control points, we can finely control the curvature, slope, and overall behavior of trajectories.The general nth-order Bezier curve equation:
\[ B(t) = \sum_{i=0}^{n} B_i^n(t) \cdot P_i \]
where \( t \) is the parameter within the range [0, 1], \( n \) is the order of the curve, \( P_i \) are the control points, and \( B_i^n(t) \) are the Bernstein basis polynomials given by:
\[ B_i^n(t) = \binom{n}{i} t^i (1 - t)^{n - i} \]
Many researchers have successfully implemented Bezier curve for trajectory generation \cite{conference:beziercurve1}\cite{conference:beziercurve2}.Some of useful properties of bezier curve are shown in \cite{book:Feedbackcontrol}. For our application, we use a fourth-order bezier polynomial. Where the first two control points correspond to the start of the trajectory and the fourth and fifth correspond to the end of the trajectory. The middle control point is set to half of the start and end leg value plus an additional user-defined value. We use this to generate trotting in-place trajectory and foot placement trajectory.

\chapter{Harpy's Simscape Results}
\label{chap: Harpy's Simscape Results}

To validate the feasibility of our trajectory and inverse kinematics concept, we constructed a Simscape model using a CAD model from Solidworks. This comprehensive model incorporated a well-designed ground contact representation, joint model, and inverse kinematic block. By implementing a precise input trajectory, we extracted data on joint torque and thruster forces, facilitating an evaluation of the input trajectory's effectiveness. This approach allowed us to gain valuable insights into the alignment between the theoretical trajectory and real-world performance.

\section{Simscape Model}

\begin{figure}[h]
  \centering
      \begin{subfigure}[b]{0.5\textwidth}
        \centering
        \includegraphics[width=0.95\textwidth]{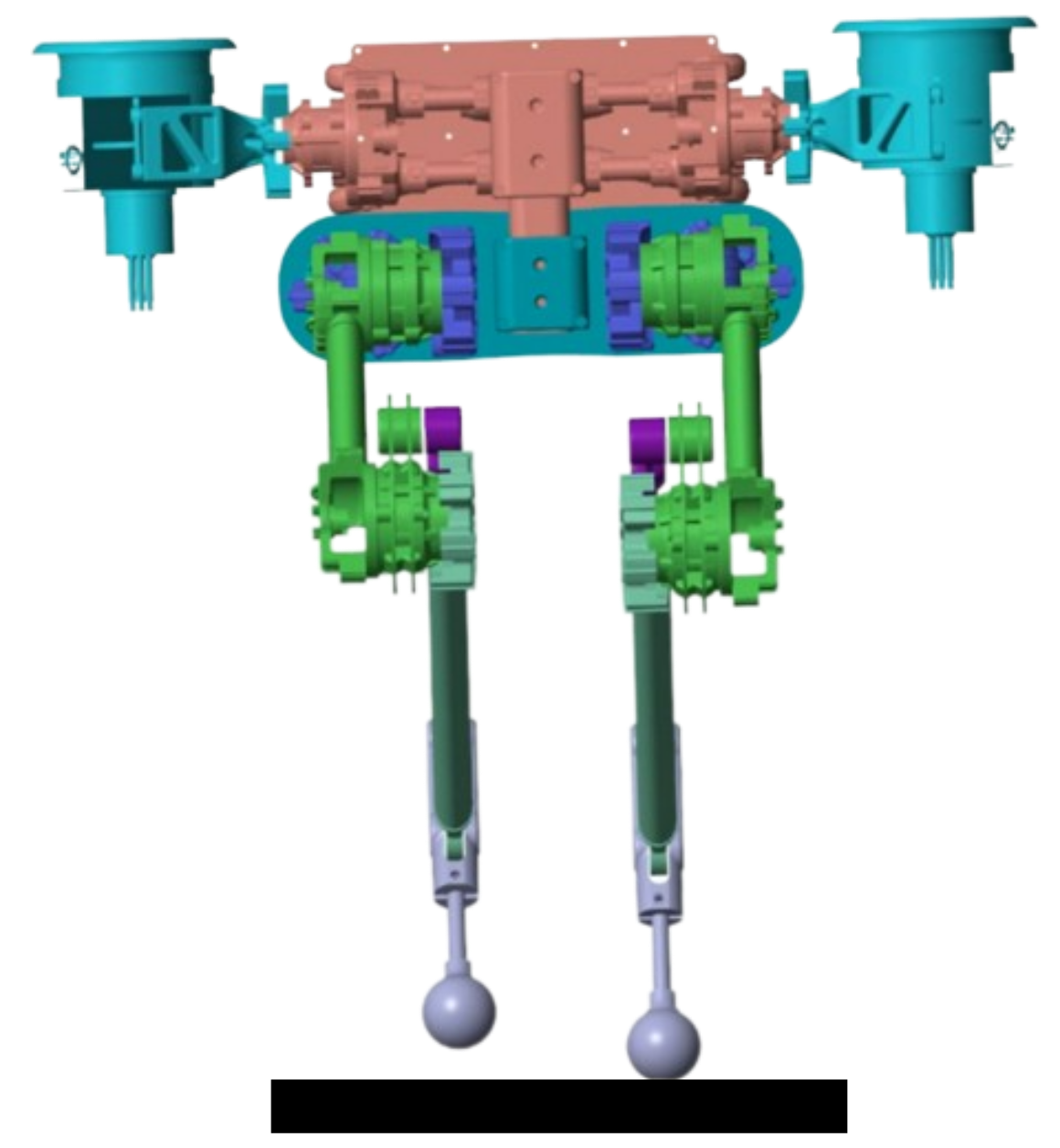}
        \caption{Front view}
        \label{fig:Simscape_front_view}
     \end{subfigure}
     \hfill
     \begin{subfigure}[b]{0.45\textwidth}
        \centering
        \includegraphics[width=0.7\textwidth]{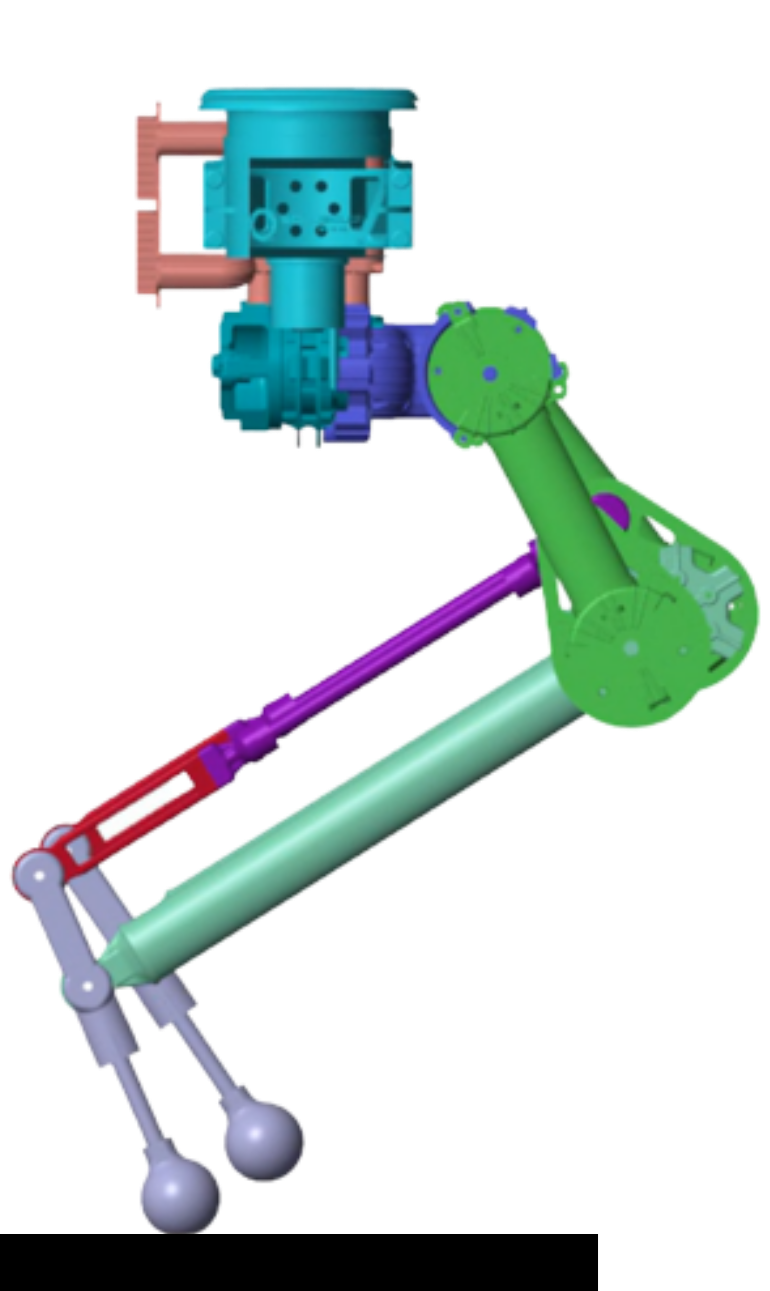}
        \caption{Side view}
        \label{fig:Simscape_side_view}
     \end{subfigure}
     \hfill
    \caption{Harpy's simscape model}
    \label{fig:HarpyComparison}
\end{figure}

The Simscape model was developed by importing the URDF model generated by Solidworks using CAD to the URDF plugin. We assigned physical attributes to the model such as mass allocation to different blocks, stiffness to joints which reflects physical condition, and foot contact parameters such as ground conditions. Within the model, Physical actuators were replicated as revolute joints. Simscapes revolute joint allows us to input joint trajectory and retrieve corresponding torque, velocity, and acceleration. Additionally, to accurately emulate shock absorption, we modeled the shock absorber as a prismatic joint, incorporating spring stiffness and damping characteristics that were sourced from relevant vendors. This comprehensive model has six revolute joints and two prismatic joints. The simscape model did not have an active foot placement controller and thus we constrained the pitch and yaw of the system. Roll and motion about XYZ cartesian coordinates were free. To calculate the thruster force while walking a PID controller was implemented. Input for the PID controller was the roll angle of the robot and output was thruster force.

To emulate the interaction between the system and the ground, we introduced the ground model into the simulation. We achieved this by creating a virtual plane and simulating contact between the foot and plane using a spatial contact force block. This specific block simulated ground interaction akin to a spring-damper system, with normal force dynamics governed by stiffness and damping parameters. The model further consists of an Inverse kinematic block which was created from the calculation shown in section \ref{Inverse kinematics}.

\begin{figure}[h]{}
        \resizebox{1.0\textwidth}{!}{\input{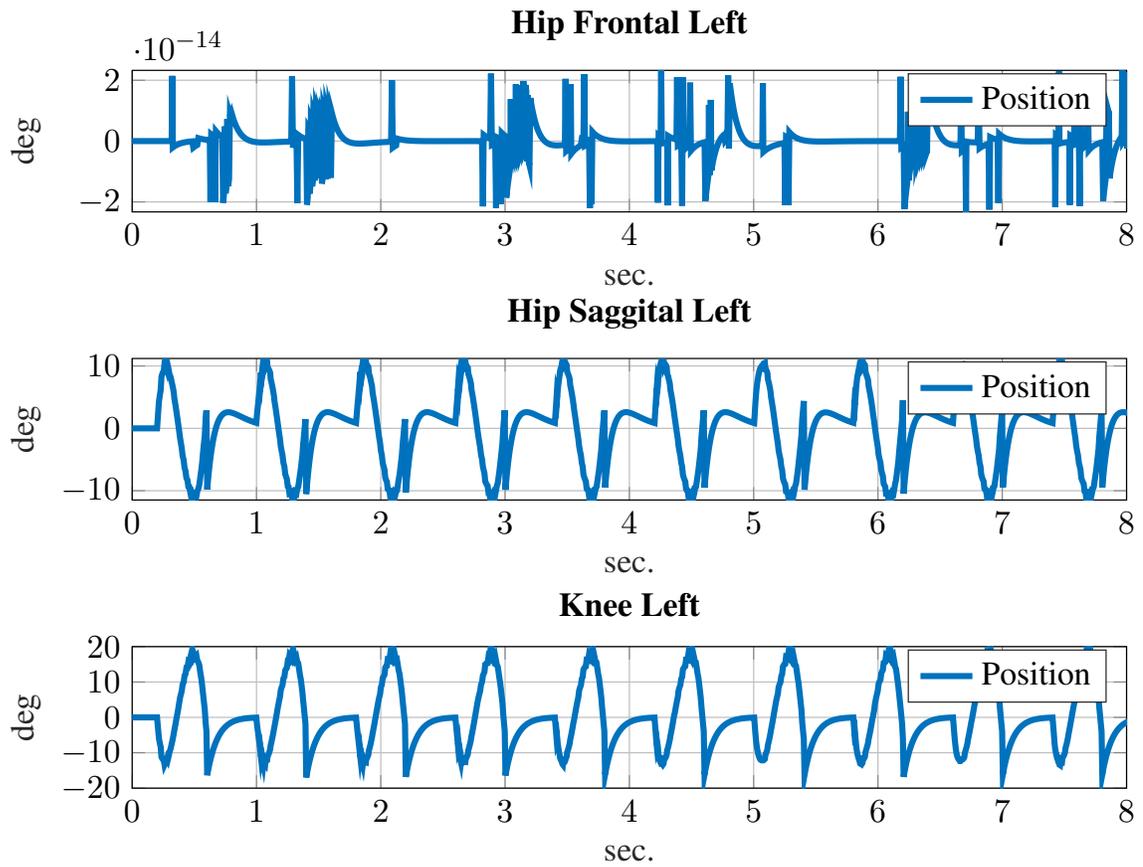}}
        \caption{Joint angles for left leg}
        \label{fig:Left leg joint angles}
\end{figure}

\begin{figure}[h]{}
        \resizebox{1.0\textwidth}{!}{\input{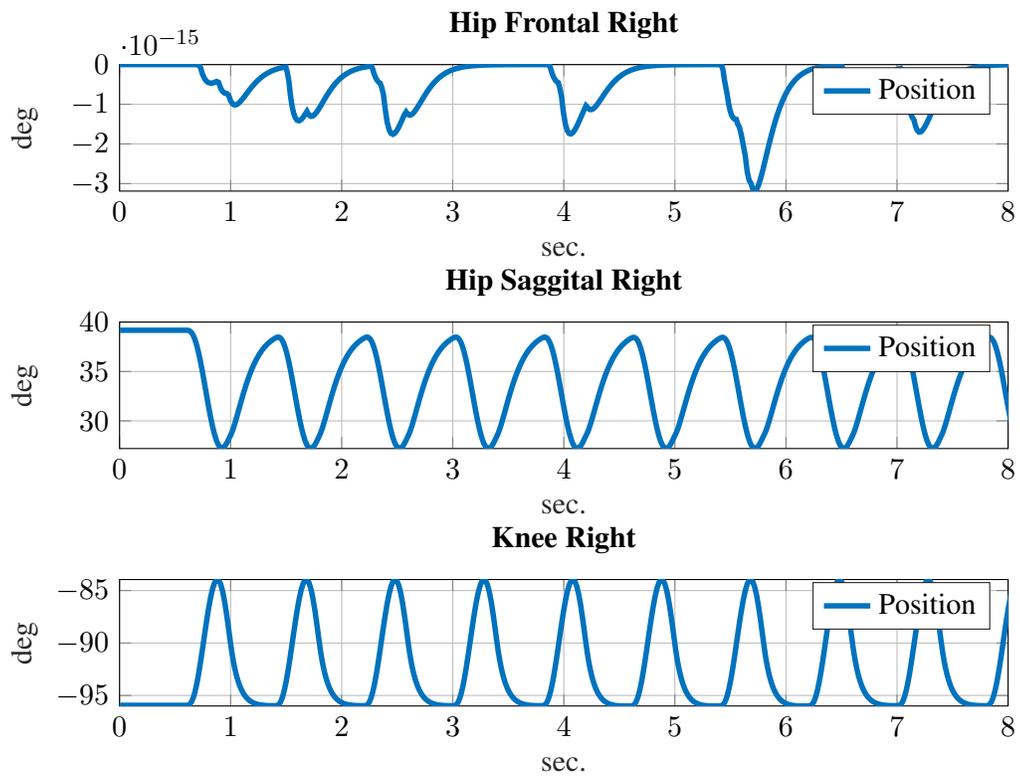}}
        \caption{Joint angles for Right leg}
        \label{fig:Right leg joint angles}
\end{figure}

\begin{figure}[h]{}
        \resizebox{1.0\textwidth}{!}{\input{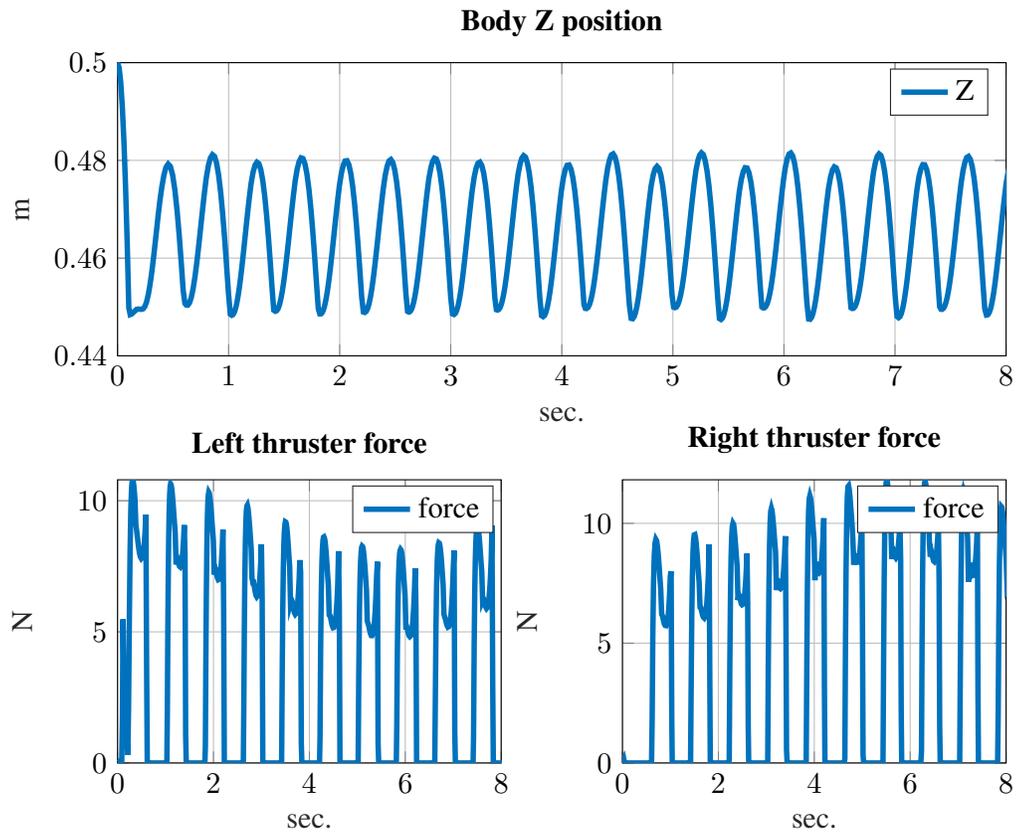}}
        \caption{Body position and thruster force}
        \label{fig:Body position}
\end{figure}

\begin{figure}[h]{}
        \resizebox{1.0\textwidth}{!}{\input{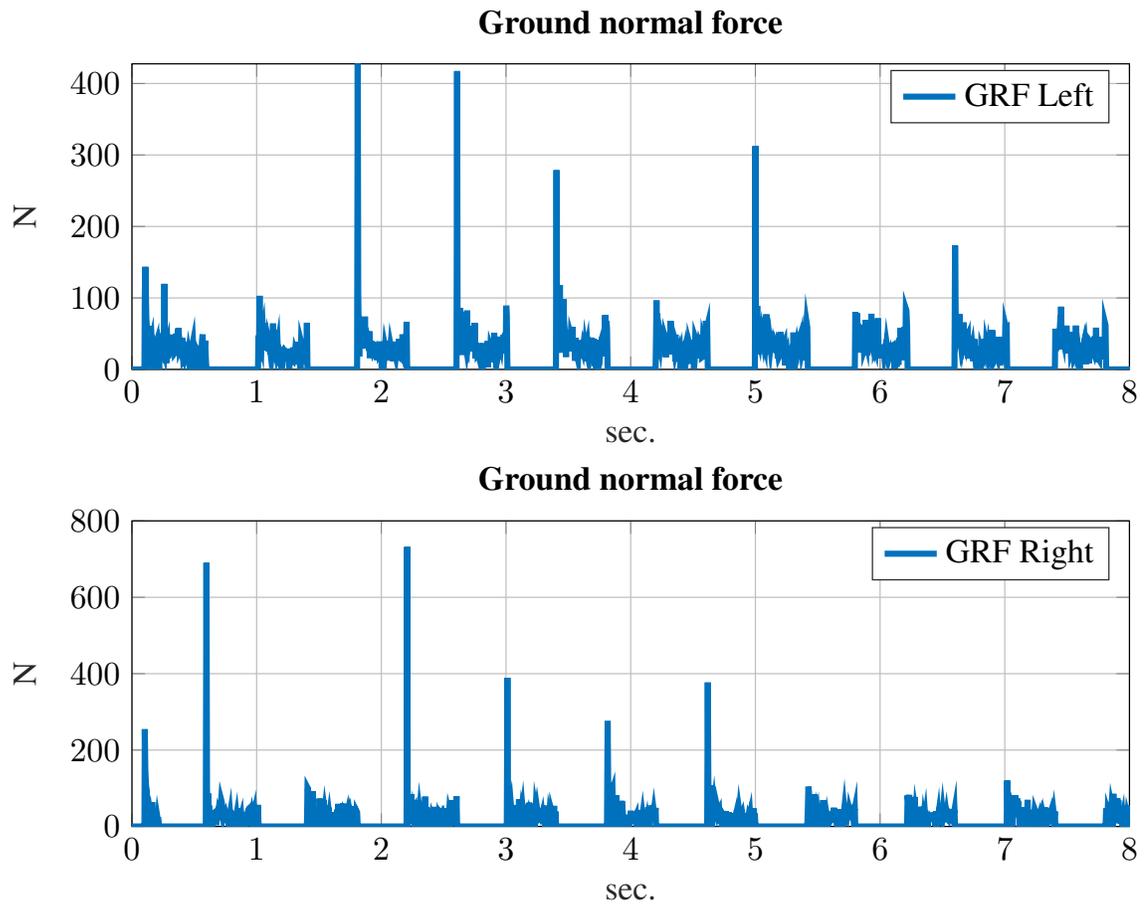}}
        \caption{Ground reaction force}
        \label{fig:GRF}
\end{figure}

We create a walking trajectory for the swing and stance phase using $4_{th}$ order bezier curve. The trajectories were implemented and simulation was performed. fig(\ref{fig:Left leg joint angles}) and fig(\ref{fig:Right leg joint angles}) shows the joint trajectories for left and right leg respectively. Harpy is walking in a straight path and thus joint angle produced for the hip frontal joint for both legs is almost $0$. From the body Z position graph in fig(\ref{fig:Body position}), we can see that the body was dropped from a height of $50$cm and it starts walking after $0.1$sec. From fig(\ref{fig:Body position}), we can see that for stable walking we would need a thruster force of about $11$N. Ground reaction force (\ref{fig:GRF}) was extracted from the Spatial contact force block.

\chapter{Testing and Results}
\label{chap:testing}
This chapter shows the experimental result of the test of trotting gait with a capture point controller to stabilize roll and pitch. To test various gaits on Harpy a testing platform was developed.

\section{Harpy's testing platform}

\begin{figure}[h]
  \centering
    \includegraphics[width=0.8\textwidth]{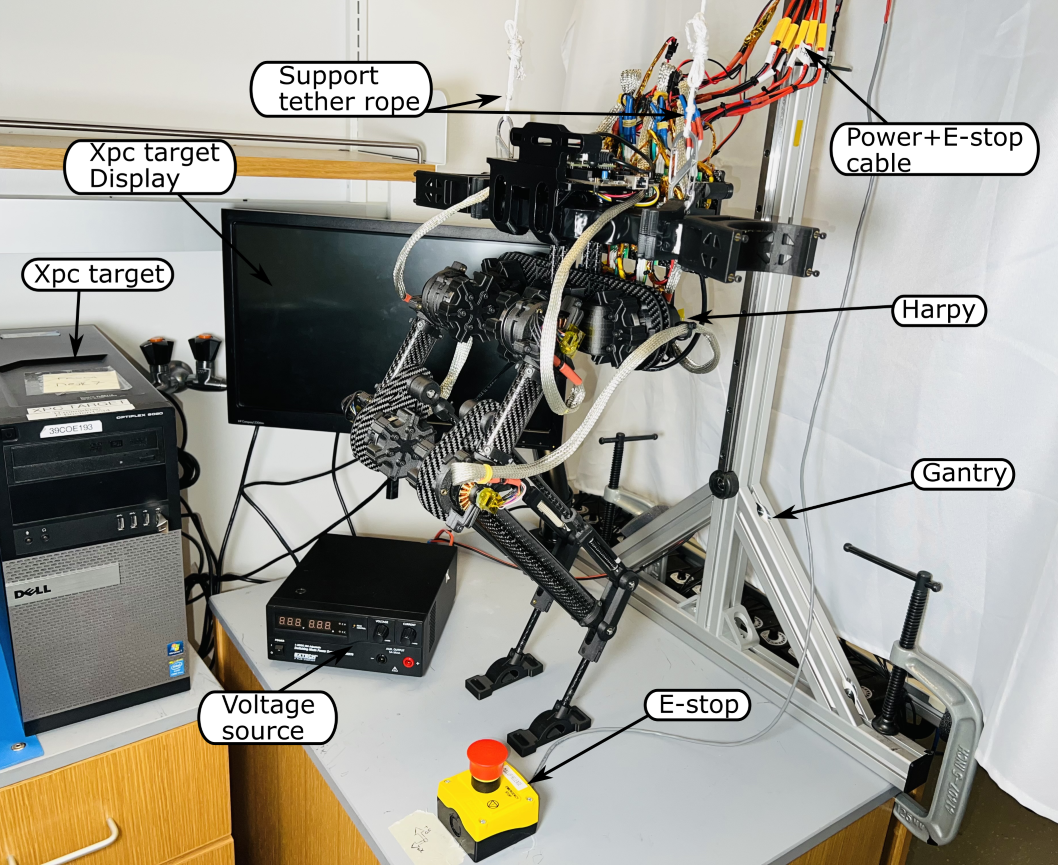}
    \caption[Harpy test fixture]{Harpy test fixture}
    \label{fig:Harpy test fixture}
\end{figure}

The purpose of the testing platform as shown in fig(\ref{fig:Harpy test fixture})is to provide support to robots with tethers and allow testing without the need for onboard systems and power supply. The arena is made from t-slot frame fixtures. The fixture is created such that there is an overhang upon which the pulleys are mounted. Tethers are connected to the thruster mount of the harpy. This setup allows us to control the load on the Harpy.

The testing arena consist of an Xpc target, external power supply, screen, and E-stop. Xpc target is a powerful tool that bridges the between simulation and real-world testing by enabling simulink models to run in real-time on dedicated hardware. The external power supply is used to power the motors and amplifier. The external power supply is an Extech 382275 600W Switching Mode DC power supply. Output is adjustable from $0$ to $30$V DC and $0$ to $20$ Amps with ($0.1$ volts/Amps resolution). For our application, we set $30$V and $14$Amps. The screen is used for the kernel display. it allows us to see input trajectories, amplifier status, and IMU data in real-time. E-stop(Emergency stop) is coupled with safe torque off (STO) unit on ELMO amplifiers which once pressed prevents the drive from producing any torque.

Harpy has its perception unit onboard. Harpy's perception unit consists of Intel RealSense tracking Camera T265 and Sparkfun ICM20948 (IMU). IMU is a versatile $9$ degree of freedom sensor module designed to provide accurate motion tracking and orientation sensing capabilities. $3$-Axis gyroscope allows FRS at $\frac{+}{-}*250$dps,$\frac{+}{-}*500$dps,$\frac{+}{-}*1000$dps and $\frac{+}{-}*2000$dps. For our application, we use $\frac{+}{-} 250$ dps. The Intel Real Sense camera includes two fisheye lens sensors.

\section{Harpy's Simulink Real-Time platform}

Harpy's simulink Real-time model was developed in MATLAB R2020a. A communication flow diagram is shown in the chapter(\ref{chap:Prototyping Harpy}). The host computer with a Simulink Real-time model is connected to the Xpc target using TCP/IP. When the Simulink real-time code is compiled, it generates C code deployed on the Xpc target. Communication with the Elmo amplifier and with IMU takes place over Ethercat.

\begin{figure}[h]
    \includegraphics[width=1\textwidth]{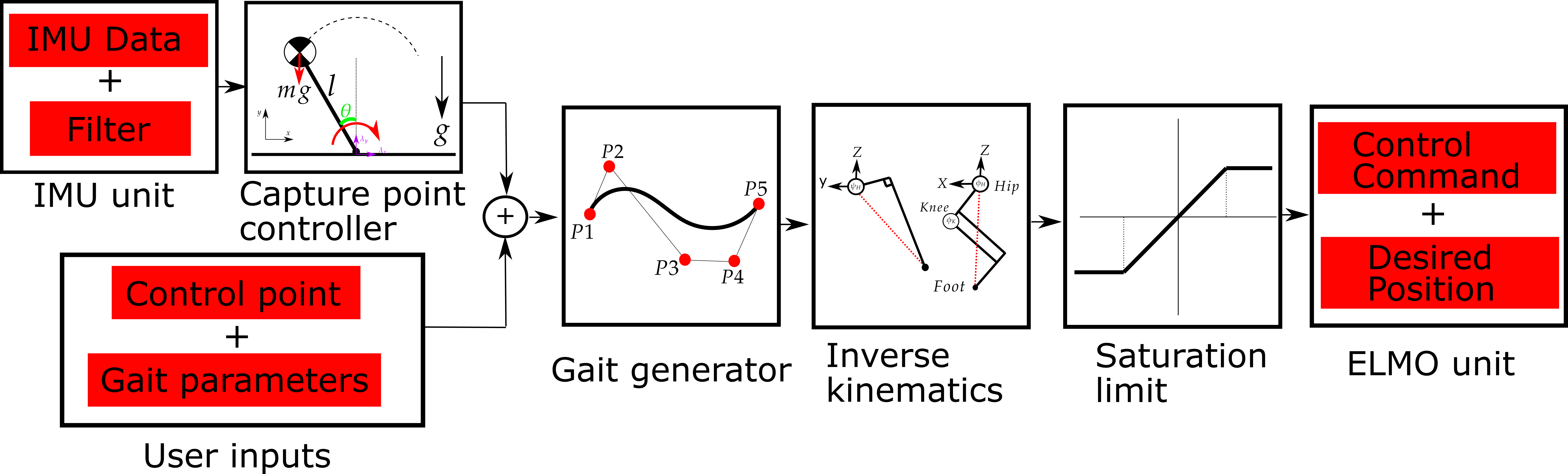}
    \caption[Simulink model]{Simulink model}
    \label{fig:Simulink model}
\end{figure}

 Figure(\ref{fig:Simulink model}) shows the flow diagram for the Simulink real-time code.The user input block is created to assign the type of gait, initial position, target position, and gait parameters like gait length, step length, and step height. IMU data obtained is passed through a low pass filter before feeding to the controller. The controller produces the foot placement value for both legs according to pitch and roll angle. Both user inputs and foot placement location are fed to the gait generator block. The gait generator block and generates a bezier curve according to the gait parameter using the five control points. Inverse kinematics block converts leg end position to joint position. Joint positions are passed through the saturation limit block to avoid the legs from colliding with themselves. Joint positions are converted to encoder positions according to encoder resolution. Finally, Encoder positions are transferred to the Elmo block which then produces the control word to command motor to move to the desired position.

\section{Trotting along with capture point generation Experiment}

\begin{figure}[h]{}
        \resizebox{1.0\textwidth}{!}{\input{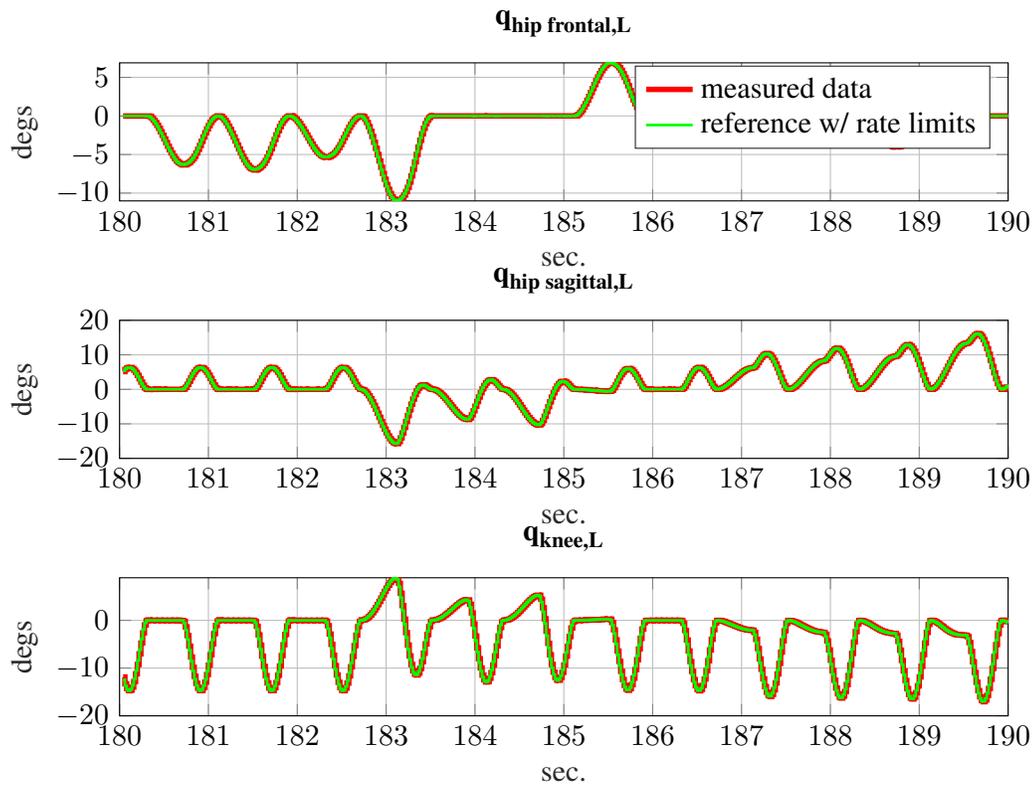}}
        \caption{Left leg joint angles}
        \label{fig:Left Joint Angle}
\end{figure}

\begin{figure}[h]{}
        \resizebox{1.0\textwidth}{!}{\input{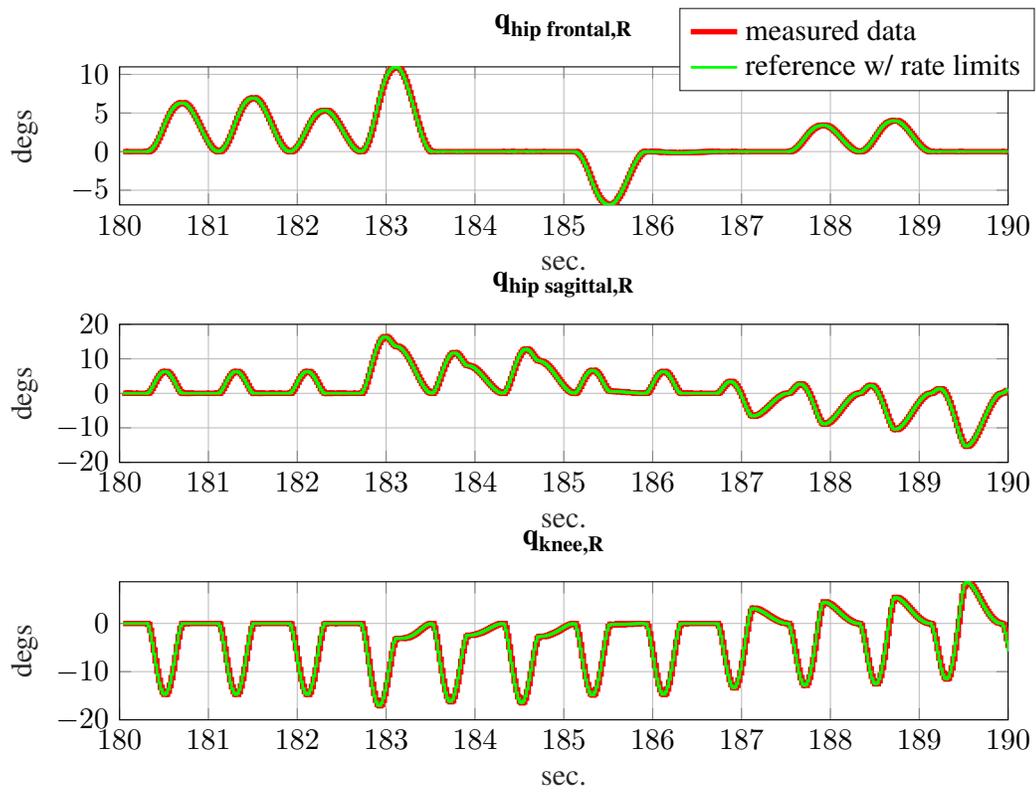}}
        \caption{Right leg joint angles}
        \label{fig:Right Joint Angle}
\end{figure}

\begin{figure}[h]{}
        \resizebox{1.0\textwidth}{!}{\input{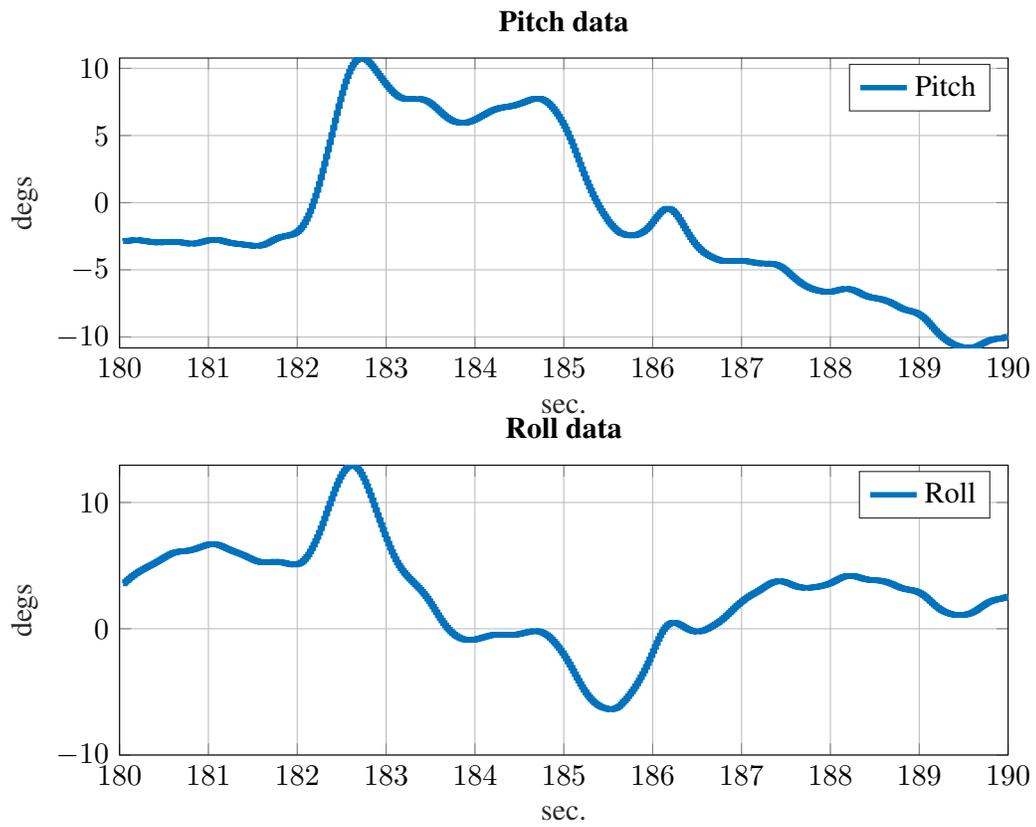}}
        \caption{ Filtered Pitch and roll data}
        \label{fig:IMU data}
\end{figure}

We performed a trotting experiment on a harpy. Harpy was supported using ropes during the initialization process. Once initialization was done, we lowered the robot and started trotting in-place gait. The capture point controller is developed in such a way that capture coordinates are only produced when pitch and roll exceed the range of ($3$ and $-3$). x-coordinate and y-coordinate are independently affected by pitch and roll respectively as seen from the plot(\ref{fig:Capture point}). The end position of the trotting gait is changed based on the captured coordinate in real-time as shown in fig(\ref{fig:Left Joint Angle}). Further, from the fig(\ref{fig:Right Joint Angle}), we can see that robot is successfully tracking the changes in position. IMU data obtained was noisy and had large spikes before settling to the correct angle. Thus, data was passed through a low-pass filter, an exponential moving average and then fed to the controller to produce capture coordinates. fig(\ref{fig:IMU data}) shows the filtered IMU data.

\begin{figure}[h]{}
        \resizebox{1.0\textwidth}{!}{\input{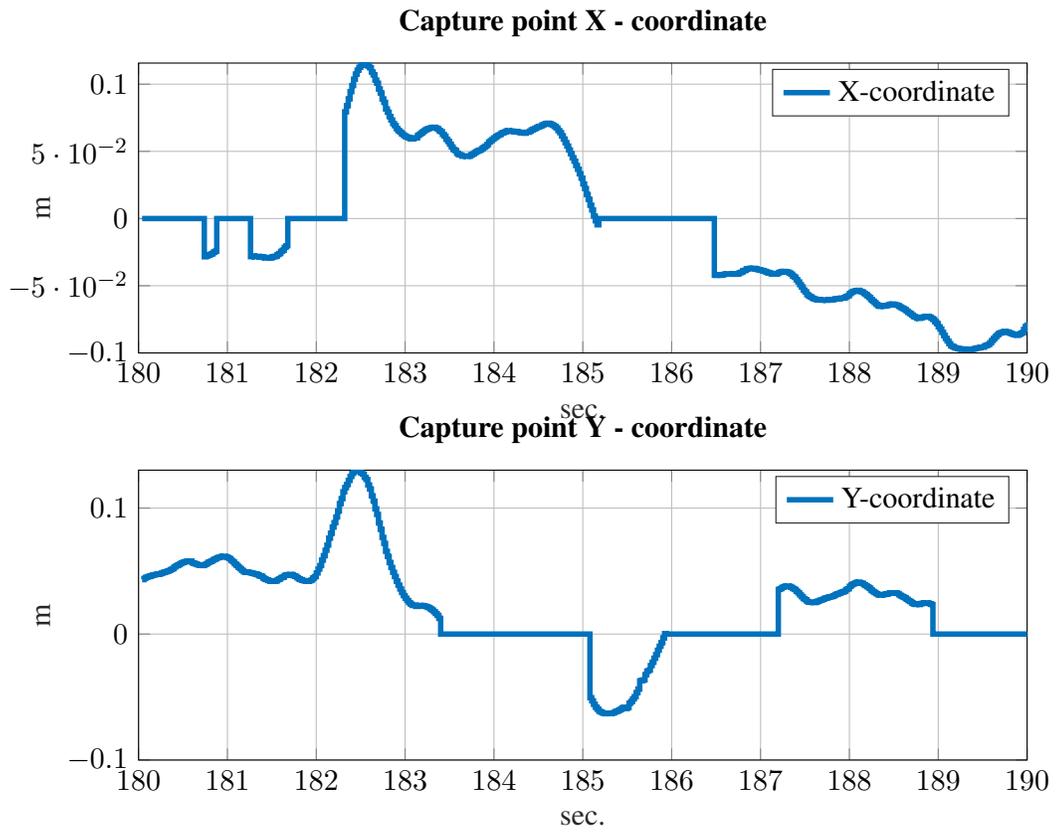}}
        \caption{Capture point co-ordinate generation based on IMU data}
        \label{fig:Capture point}
\end{figure}

 \chapter{Conclusion}
\label{chap:conclusion}

The outcome of my thesis defines the robustness of the hardware and effectiveness of the controller which allows harpy to stabilize itself. In this thesis, I have introduced the goals of the Harpy project and its innovative design which was inspired by Widowbird. Harpy's legged design was built upon the husky generative design\cite{paper:Huskygenerative}. Due to Harpy's thruster, it has an advantage over the traditional bipedal robot and can perform complex tasks with ease. Next, Harpy's fabrication is shown where we delve deep into prototyping Harpy's actuator, Pelvis assembly, Ankle assembly, and thruster assembly. Most of the components of the Actuator, pelvis assembly, and ankle assembly are 3D printed and further reinforced with carbon fiber using the concept of sandwich panel theory. Components like bearing and heat-insert are embedded inside the housing to reduce the overall usage of fasteners. After fabrication, Actuators were tuned and tested using Elmo Studio software to verify their performance capability. The thruster mount was created using a carbon fiber-aluminum composite structure to achieve a lightweight and higher strength-to-weight ratio. Using inverse kinematics, we found the joint angles for a particular foot end location and developed the MATLAB script from equations. There are various challenges in controlling the bipedal robot including dynamics stability, gait design, and sensor feedback. Early attempts at bipedal robot control often struggled with replicating human-like agility and robustness. The dynamic nature of bipedal walking, with its need for constant balance adjustments, led researchers to explore new control paradigms. Amidst these challenges, a significant breakthrough emerged with the development of the Capture Point Theory. We model Harpy's walking dynamics as a 3D linear inverted pendulum model and apply capture point theory. Capture point theory using orbital energy concept as it remains constant during locomotion. Furthermore, by equating orbital energy to zero, we can find the capture point. Using the capture point equation, we developed a controller that takes input as the Pitch and roll angle of the robot and provides the next foot location for the Harpy to stabilize itself. 

The simscape model was developed for testing the inverse kinematics model and to find the thruster force required for a stable walking gait. The walking trajectory was developed using $4_{th}$ order bezier curve and implemented. Using the simscape model, I was able to verify that inverse kinematic calculations were correct and that for stable walking, we need $12$N of individual thrust force. Harpy test gantry was created in such a way that we can change the amount of load taken by Harpy. The Simulink model was created which consists of capture point control, IK block, Elmo amplifier block. The Simulink model was developed and deployed in Xpc target which is connected to hardware and testing was performed. From the test, I was able to verify the effectiveness of the capture point controller. As the controller was tracking in the changes in pitch and roll effectively. 

\section{future work}

With the finalization of the thruster, next is installing the thruster on the robot. Performing thruster-assisted walking tests on the hardware and comparing results against the simulated results shown in my thesis. Further, extend capture point theory by adding thruster dynamics and creating a control strategy out of it. Moreover, implementing ERG optimization on hardware to manipulate the ground reaction force \cite{unknown:GRFHarpy}

\bibliographystyle{IEEEtran}  

\bibliography{bib/thesis}



\end{document}